\documentclass[default,pdflatex,iicol]{sn-jnl} 

\usepackage[caption=false,font=normalsize,labelfont=sf,textfont=sf]{subfig}
\usepackage{todo}
\usepackage{textcomp}
\usepackage{stfloats}
\usepackage{url}
\usepackage{verbatim}
\usepackage{graphicx}
\usepackage{multirow}
\usepackage{rotating}
\usepackage{longtable}
\usepackage{afterpage}
\usepackage{multicol}
\usepackage[english]{babel}
\usepackage{natbib}
\usepackage{array}
\usepackage{tabto}
\NumTabs{8}

\newcommand{\quoted}[1]{``#1''}

\newcommand{\framework}{ODIN }
\newcommand{\ntools}{29 }
\newcommand{\nmetrics}{40 }

\newcolumntype{K}[1]{>{\centering\arraybackslash}p{#1}}

\begin{document}

\title[Black-box Error Diagnosis in Deep Neural Networks for Computer Vision: a Survey of Tools]{Black-box Error Diagnosis in Deep Neural Networks for Computer Vision: a Survey of Tools}

\author[1]{\fnm{Piero} \sur{Fraternali}}\email{piero.fraternali@polimi.it}
\equalcont{Authors are listed alphabetically.}

\author*[1]{\fnm{Federico} \sur{Milani}}\email{federico.milani@polimi.it}
\equalcont{These authors contributed equally to this work.}

\author[1]{\fnm{Rocio Nahime} \sur{Torres}}\email{rocionahime.torres@polimi.it}
\equalcont{These authors contributed equally to this work.}

\author[1]{\fnm{Niccolò} \sur{Zangrando}}\email{niccolo.zangrando@polimi.it}
\equalcont{These authors contributed equally to this work.}

\affil[1]{\orgdiv{Department of Electronics, Information and Bioengineering (DEIB)}, \orgname{Politecnico di Milano}, \orgaddress{\street{Via Giuseppe Ponzio, 34}, \city{Milan}, \postcode{20133}, \state{MI}, \country{Italy}}}

\abstract{
The application of Deep Neural Networks (DNNs) to a broad variety of tasks demands methods for coping with the complex and opaque nature of these architectures.  When a gold standard is available,  performance assessment treats the DNN as a black box and computes standard metrics based on the comparison of the predictions with the ground truth. A deeper understanding of performances requires going beyond such evaluation metrics to diagnose the  model behavior and the prediction errors. This goal can be pursued in two complementary ways. On one side, model interpretation techniques  \quoted{open the box} and assess the relationship between the input, the inner layers and the output, so as to identify the architecture modules most likely to cause the performance loss. On the other hand,  black-box error diagnosis techniques study the correlation  between  the model response and some  properties of the input not used for training, so as to identify the features of the inputs that make the model fail. Both approaches give hints on how to improve the architecture and/or the training process.
This paper focuses on the application of DNNs to Computer Vision (CV) tasks and presents a survey of the tools that  support the  black-box performance diagnosis paradigm. It  illustrates the features and gaps of the current proposals, discusses  the relevant research directions and  provides a brief overview of the diagnosis tools in  sectors other than CV.
}

\keywords{black-box, error diagnosis, machine learning, evaluation, metrics}

\maketitle

\section{Introduction}

The advances in Machine Learning (ML) led to the development of more complex neural networks nowadays known as Deep Neural Networks (DNNs), which are powerful models able to process complex data of various types \cite{liu2017survey}. DNNs have achieved outstanding results in many diverse domains and have become the solution of choice for addressing big data analysis  \citep{chiroma2018progress}. One of the domains in which DNNs have attained the most impressive results is Computer Vision (CV) where they have outperformed previous methods in a variety of tasks, including image and scene classification, object detection, semantic and instance segmentation, object and activity tracking and pose estimation   \citep{voulodimos2018deep}.

The  life cycle of a DNN application differs from the  development workflow of a traditional software system, because it relies on predefined algorithms that must be trained for the specific task and data  at hand  \citep{gharibi2021automated}.

\begin{figure}
    \centering
    \includegraphics[width=0.9\linewidth]{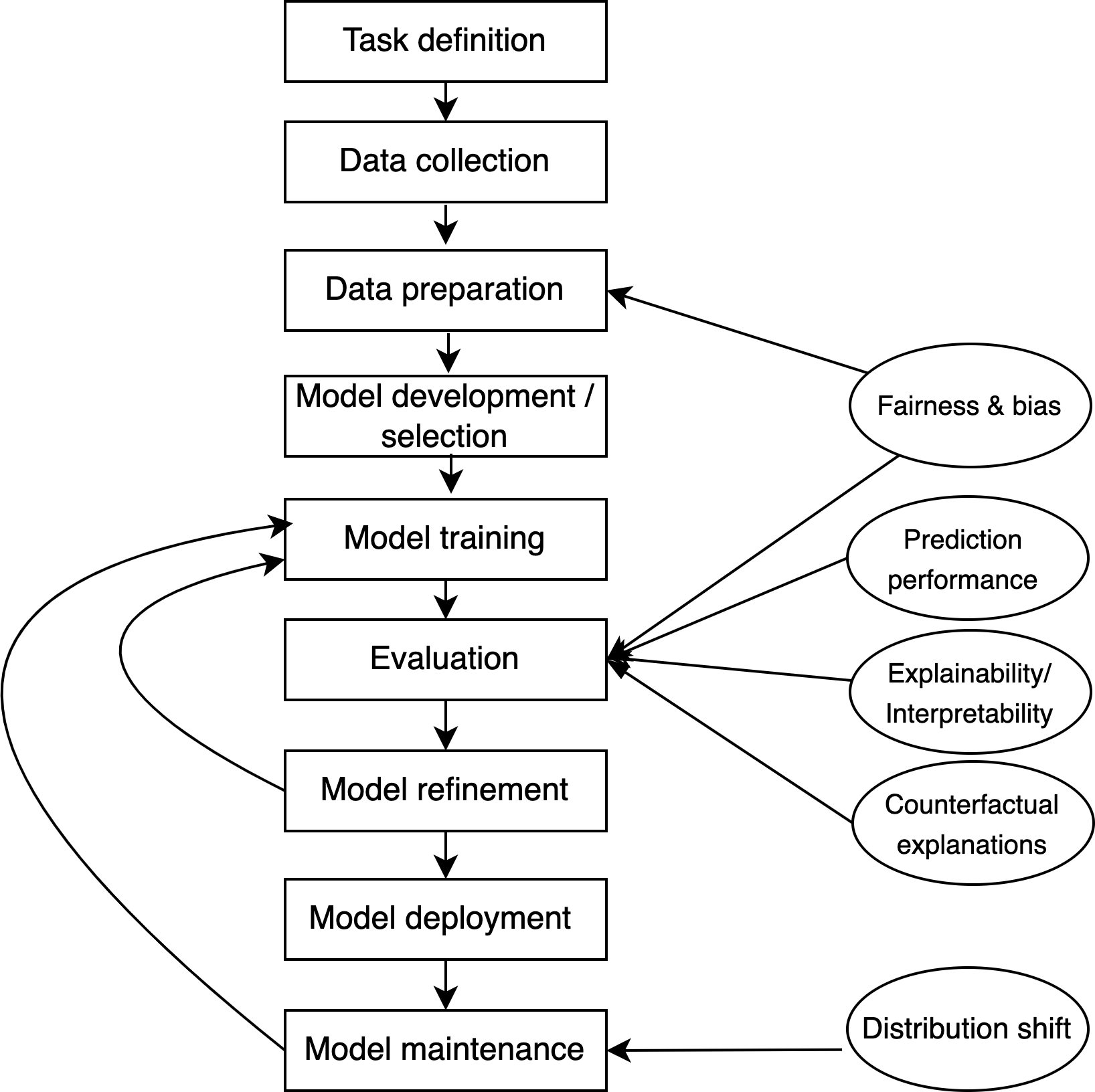}
    \caption{Life cycle of a DNN application}
    \label{fig:MLdevproc.drawio}
\end{figure}

Figure \ref{fig:MLdevproc.drawio} illustrates the typical development workflow of a DNN-powered application. 

The peculiarity of such a workflow is the selection of the predefined algorithm more suited to the task and its fitting to the problem data via training, which introduces a  distinction between the preparatory stage and the execution (inference) stage. 
A further specificity of DNNs  is their complex and opaque nature, which makes debugging particularly hard. 

The ordinary way of assessing DNNs is the evaluation of their prediction performance with a gold standard.  Prediction performance analysis treats the DNN as a black-box and compares  its output with the ground truth with metrics such as accuracy, precision, recall, etc.
As DNNs get more and more applied to critical tasks, the need arises for  a deeper understanding of the way in which predictions are made. This insight can be exploited  to justify the output and/or to improve the performances.

The investigation of the behavior of ML models in general, and of DNNs in particular,  can be pursued with analysis approaches that focus on distinct aspects:

\begin{itemize}
    \item Prediction performance: is the assessment of the quality of the predictions.  Performance can be evaluated either qualitatively via manual inspection or quantitatively by comparison with ground truth test data. Quantitative performance analysis exploits standard metrics, such as Accuracy, Precision, Recall, F1-Score, or Average Precision.
    \item Model interpretability and explainability: is the ability to explain or to present in understandable terms to a human how a model makes a prediction  \citep{doshi2017towards,GuidottiMRTGP19}.  Intrinsic interpretability refers to those models that are simple and thus interpretable by design, such as short decision trees or sparse linear models. Conversely, post-hoc interpretability requires specific investigation techniques applied to the trained model. In DNNs, post hoc interpretability  aims at exposing the relation of the internal representations of deep models to the input and output  \citep{zhang2018visual,MONTAVON20181,Carvalho2019,TjoaG21, barredo2021post}. Techniques such as the Class Activation Maps (CAMs)  \citep{zhou2015cnnlocalization,Grad-CAM2017,Chattopadhay_2018,Sun2020,BaeNK20} highlight the most influential regions of the feature maps at different network levels and  enable the insight into  the model prediction process. The difference between interpretability and explainability is subtle: the former emphasizes human intuition whereas the latter stresses the comprehension of the internal logic of the model  \citep{linardatos2020explainable}.
    \item Counterfactual explanations: is the approach that exposes the model behavior by showing how some actions, such as a change to the input, alter the behavior of the system   \citep{verma2020counterfactual, stepin2021survey}. 
    \item Fairness and bias: is the assessment of how an algorithm delivers  predictions when applied to inputs belonging to  data populations with different characteristics (e.g., to images of people belonging to different demographic groups)  \citep{mehrabi2021survey}.
    \item Distribution shift:  is the evaluation of how the model performances evolve when the distribution of data changes with respect to the one of the training and testing data  \citep{wu2021methods}.
\end{itemize}

The support of computer tools to the development life cycle of DNN applications has concentrated mostly on the basic tasks of model development/selection, training and testing. Several software packages provide off-the-shelf functionalities for defining the structure of a DNN, executing the training process and evaluating the most common metrics on the test set  \citep{wang2019various}. As the application of DNNs matures and becomes a common practice in the industry, tool support beyond model definition, training and testing needs to be developed. This paper  surveys the status of a specific sector of the research in this direction: black-box error diagnosis tools for CV tasks.

\subsection{Focus of the survey}

The survey has a threefold focus: on the application task,  on the  analysis type, and on the life cycle phase.

The \textbf{task focus} concentrates on the use of DNNs for CV. The motivation for this choice is the fact that CV is the sector in which the availability of black-box diagnosis tools is more significant. To complete the overview, we also provide in Section \ref{sec:beyond} a brief appraisal of the status of  DNN diagnosis tools in other domains, such as time series analysis,   natural language processing and recommender systems.

The \textbf{analysis focus} concentrates on prediction performance as the target of black-box diagnosis. Among the analysis approaches mentioned above,  explainability and interpretability have already been described in several  survey papers  \citep{gilpin2018explaining, choo2018visual, roscher2020explainable, molnar2020interpretable}.   Fairness and bias detection is also well documented in several overview works  \citep{mehrabi2021survey, pessach2022review}, as well as distribution shift  \citep{wu2021methods}. Counterfactual explanations have been long used in statistical learning and have been recently rediscovered as a DNN explainability technique: the works  \citep{verma2020counterfactual, stepin2021survey} document the status of research in that area.

The \textbf{life cycle} focus considers the tools that offer computerized support to the model evaluation and refinement steps.

When appropriate, if a tool  supporting primarily the black-box  diagnosis approach offers also functionalities for other types of analysis and/or development steps, these will be mentioned too.

\subsection{Methodology}

The target of the research comprises those methods that exploit only knowledge about the input and output to compute and break down performance metrics and to characterize errors. Among such works, we highlight the proposals that provide a tool for DNN evaluation, possibly together with functions for model design and training. This perimeter excludes those contributions that address DNN behavior and performances but pursue different targets such as special-purpose and  domain-dependent  metrics, the visualization of DNN internal representations  \citep{roscher2020explainable}, model design for interpretability  \citep{doshi2017towards}, and  human-in-the-loop interpretation  \citep{BalaynSL0B21}. 

\begin{figure}
    \centering
    \includegraphics[width=0.81\linewidth]{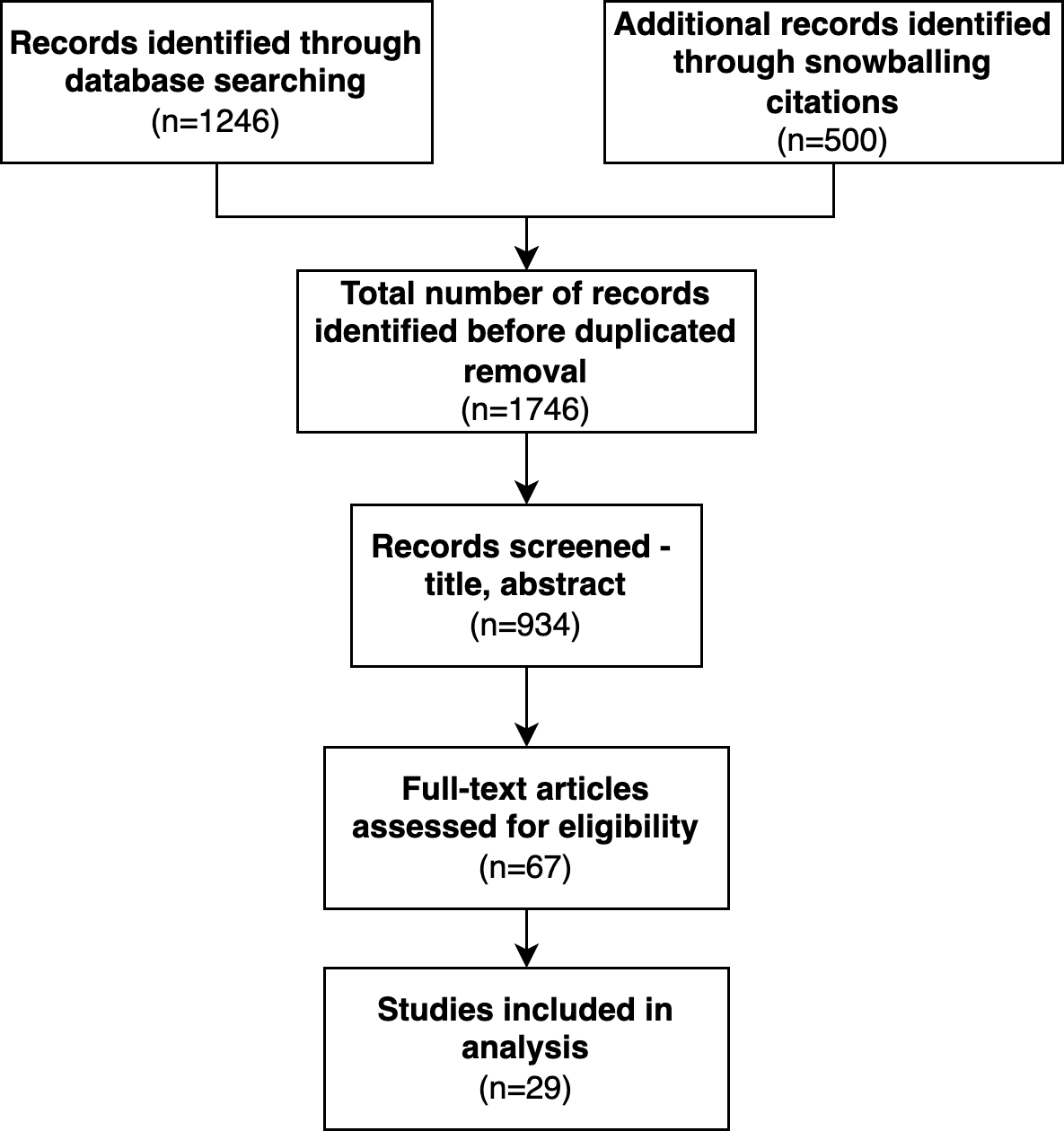}
    \caption{PRISMA flow diagram of the  systematic review}
    \label{fig:prismasearch}
\end{figure}

The corpus of the relevant research has been identified by  following a simplified PRISMA procedure   \citep{page2021prisma} for systematic reviews.      Figure \ref{fig:prismasearch} illustrates the adopted workflow. 

\begin{enumerate}
    \item  The search has been conducted on the Scopus  repository, since previous studies  have shown that it  supports bibliographic research better than other sources \citep{falagas2008comparison}. The  search phrases have been composed as follows: 
\begin{small}
\begin{verbatim} 
<search> :- <task> AND <goal> AND <system>
<task> :- machine learning | deep learning | 
          computer vision | classification | 
          image classification | 
          scene classification | 
          object detection |  
          instance segmentation | 
          semantic segmentation | 
          object tracking | 
          pose estimation |  
          activity detection | 
          action detection 
<goal> :- model diagnosis | 
          error diagnosis | 
          performance analysis
<system> :- tool | framework | workbench
\end{verbatim}
\end{small} 
The output of the search was filtered to retain only contributions in journals, conferences and workshops.
  \item The initial corpus has been expanded through snowballing, recursively adding further related studies citing or cited by the  works in the initial corpus.
  
  \item The expanded corpus, composed of 1,746 works, has been reduced by removing duplicates. Next, we have identified and eliminated the studies unrelated to black-box techniques, by checking the title, keywords and  abstract of each contribution. The reduced corpus contained 67 contributions.
  
  \item A final eligibility filter has been applied by reading the full-text of the  articles in the reduced corpus. This final step yielded the \ntools works considered in this survey.
\end{enumerate}

\subsection{Contributions}
The contributions of  this paper can be summarized as follows:

\begin{itemize}
    \item \ntools black-box DNN performance diagnosis  tools are identified  from a initial corpus of 1,746 papers resulting from keyword and cross-references search.
    \item The tools are described and compared based on different dimensions (year, task, data type, metrics, performance and error break down functions, openness and extensibility)
    \item A list of open issues and relevant research directions are identified and discussed.
    \item A brief overview of the status of the DNN diagnosis tools in sectors other than CV is provided.
\end{itemize}

The rest of the paper is organized as follows:  Section \ref{sec:tool-dims} describes the dimensions used to categorize the surveyed tools,  Section \ref{sec:tool-desc} and \ref{sec:tool-cmp} describe and compare the different tools based on the identified dimensions; Section \ref{sec:issues} highlights the  open issues and discusses the relevant research directions; finally, Section \ref{sec:concl} draws the conclusions.

 \section{Classification of the  tools}\label{sec:tool-dims}

The relevant proposals are described and compared along six dimensions: task, media types, metrics, break down functions, custom properties, and openness and extensibility.

\textbf{Task:} An error diagnosis tool is typically designed for performance analysis of a specific task, which in turn may apply to a specific media type or to a range of media types. The surveyed diagnosis tools span the following tasks:
\begin{itemize}
    \item Classification (CL): assignment of class labels to  input samples. The samples used in CV applications belong to visual media; however, a tool may apply the same classification performance diagnosis functions to other data types, such as text, aural content, and record data. Therefore we consider classification in general and not only image classification. 
    \item Object Detection (OD): localization of objects of a certain class through bounding boxes in images and videos.
    \item Semantic Segmentation (SS): assignment of a class label to each pixel in images or videos.
    \item Instance Segmentation (IS): similar to semantic segmentation but multiple objects of the same class are treated as separate instances.
    \item Object Tracking (OT): similar to object detection but each unique object is tracked as it moves across the frames of a video.
    \item Pose Estimation (PE): recognition of single or multiple body poses through key points in image and video.
    \item Action Detection (AD): assignment of an action label to a video.
    \item Video Relation Detection (VRD):  spatio-temporal localization of object and subject pairs in videos and assignment of a label that describes their interaction.
\end{itemize}

\textbf{Media types:} Depending on the task, several media types can be relevant. The media types processed by the surveyed tools comprise image and video. The \quoted{generic} media type is used to refer to values of arbitrary record type. 
 
\textbf{Metrics:} The quantitative analysis relies on metrics that may vary based on the targeted task. Near \nmetrics metrics are mentioned in the surveyed tools. The definition of the non standard metrics can be found in the publications cited in the comparison tables.

\textbf{Break down functions:} In addition to the computation of the performance metrics, some tools implement functions that improve the characterization of the input and of the predictions, by breaking down input data sets, metrics and  errors based on several criteria. 
\begin{itemize}
    \item Overall / Per-class / Per-property performance analysis: the tool supports the computation of the metrics for the entire data set and/or for individual classes or properties of the input.
    \item Overall / Per-class / Per-property reporting: the tool supports the construction of summary reports for the different levels of granularity used to compute the performance  metrics.
    \item Categorization of errors: the tool supports the attribution of errors to  specific categories, e.g., confusion with similar/dissimilar classes, poor localization, occlusion, etc.
    \item Error contribution isolation: the tool supports error impact analysis by highlighting the effect of all errors of a certain type on the performance metrics.
    \item Properties / Class distribution: the tool visualizes the distribution of properties and  classes of the input data set.
\end{itemize}

\textbf{Custom property editing:} Some tools integrate a framework for adding custom properties to the input samples, with the following features: 
\begin{itemize}
    \item Manual annotation creation: a graphical interface allows the user to add annotations to the input samples.
    \item Annotation purpose selection: annotation can be distinguished into those for training (e.g., class labels) and those for diagnosis (e.g., domain dependent  properties).
    \item Automatic annotation extraction: the tool enables the execution of algorithms for extracting meta-data and associating them  to the input samples as custom properties.
    \item Selective visualization: the user can display the input data set and its annotations with multiple criteria (e.g., all samples of a certain class or with a specific property).
\end{itemize}

\textbf{Openness and extensibility:}
Openness and extensibility are fundamental properties to support adoption especially when novel metrics or diagnosis approaches are proposed. The surveyed tools have been assessed based on their open source status and on the effort required for their extension. This qualitative dimension is characterized by means of the following values:
\begin{itemize}
    \item Open source: the code is public and freely available.
    \item User-defined metrics: the tool can be extended with custom metrics without modifying the framework.
    \item Data set independence: the tool can be applied to   multiple data sets.
    \item User-defined properties: the tool enables the plug-in of modules that implement custom analysis types not present in the original proposal.
\end{itemize}

\section{Tool characterization and description}\label{sec:tool-desc}

Table \ref{tab:tools} lists the \ntools identified tools with the name of the tool or of its authors, the publication year, the targeted tasks, the  media types, the ability to work with different data sets, and the link to the source code (only for open-source tools)\footnote{The link to the code repository is navigable in the online version of the paper.}.

\begin{table}[h!]
\caption{The surveyed tools listed by ascending year of publication. In the \textit{Code} column, \quoted{-} indicates that the code is not available and \quoted{link} contains a reference to the code repository}
\resizebox{\columnwidth}{!}{
\centering
\begin{tabular}{p{2.1cm} c p{0.8cm} p{0.8cm} c c }
\hline
 \multicolumn{1}{c}{\textbf{Reference}} &
  \textbf{\textbf{Year}} &
  \textbf{\textbf{Task}} &
  \textbf{\textbf{Media}} &
  \textbf{\begin{tabular}[c]{@{}c@{}}Data set\\ indepen-\\dence\end{tabular}}  &
  \multicolumn{1}{c }{\textbf{Code}} \\ \hline
Dollar et al.\cite{dollar2009pedestrian} & 2009 & OD & image & no & - \\ \hline

Hoiem et al.
\cite{hoiem2012diagnosing} & 2012 & OD & image & yes & \href{https://github.com/wk910930/diagnosing-object-detectors}{link} \\ \hline

Russakovsky et al.
\cite{russakovsky2013detecting} & 2013 & OD & image & no & - \\ \hline

 \begin{tabular}[l]{@{}l@{}}COCO API \\ \cite{mscoco}\end{tabular}  &
2014 & OD, IS, PE & image & yes & \href{https://github.com/cocodataset/cocoapi}{link} \\ \hline

Hariharan et al.
\cite{hariharan2014simultaneous} & 2014 & IS & image & no & \href{https://github.com/bharath272/sds\_eccv2014}{link} \\ \hline

Zhu et. al.
 \cite{zhu2015diagnosing} & 2015 & OD & image & no & - \\ \hline
 ModelTracker \cite{amershi2015modeltracker} & 2015 & CL & generic & yes & - \\ \hline

Redondo et al.
 \cite{redondo2016pose} & 2016 & PE, OD+PE & image & no & \href{https://github.com/gramuah/pose-errors}{link} \\ \hline

 Prospector \cite{krause2016interacting} & 2016 & CL & generic & yes & - \\ \hline

Zhang et al.
 \cite{zhang2016far} & 2016 & OD & image & no & - \\ \hline

Ronchi et al.
 \cite{ruggero2017benchmarking} & 2017 & PE & image & yes & \href{https://github.com/matteorr/coco-analyze}{link} \\ \hline

 Explanation Explorer \cite{krause2017workflow} & 2017 & CL & generic & yes & \href{https://github.com/nyuvis/explanation\_explorer}{link} \\ \hline

 Squares \cite{ren2016squares} & 2017 & CL & generic & yes & - \\ \hline

Sigurdsson et al.
\cite{sigurdsson2017actions} & 2017 & AD & video & no & \href{https://github.com/gsig/actions-for-actions}{link} \\ \hline
 
 \begin{tabular}[l]{@{}l@{}}DETAD \\ \cite{alwassel2018diagnosing}\end{tabular} 
& 2018 & AD & video & yes & \href{https://github.com/HumamAlwassel/DETAD}{link} \\ \hline

Nekrasov et al.
 \cite{nekrasov2018diagnostics} & 2018 & SS & image & no & - \\ \hline

 \begin{tabular}[l]{@{}l@{}}Manifold \\ \cite{zhang2018manifold}\end{tabular} 
& 2018 & CL & generic & yes & \href{https://github.com/uber/manifold}{link} \\ \hline

 What If Tool \cite{wexler2019if} & 2019 & CL, BD & generic & yes & \href{https://github.com/pair-code/what-if-tool}{link} \\ \hline

 TIDE \cite{bolya2020tide} & 2020 & OD, IS & image & yes & \href{https://github.com/dbolya/tide}{link} \\ \hline

 ODIN \cite{torres2020odin, torres2021odin} & 2020 & OD, IS, CL & \begin{tabular}[c]{@{}c@{}}image/\\ generic\end{tabular} & yes & \href{https://github.com/rnt-pmi/odin}{link} \\ \hline

Padilla et al.
\cite{padilla2020survey} & 2020 & OD & image & yes & \href{https://github.com/rafaelpadilla/Object-Detection-metrics}{link} \\ \hline

 TF-GraF \cite{yoon2020tensorflow} & 2020 & OD & image & yes & \href{https://github.com/boguss1225/ObjectDetectionGUI}{link} \\ \hline

 Boxer \cite{gleicher2020boxer} & 2020 & CL & generic & yes & - \\ \hline
 OpenVINO \cite{demidovskij2021openvino} & 2020 & CL, OD, SS, IS & images & yes & - \\ \hline

Padilla  et al.
\cite{padilla2021comparative} & 2021 & OD, OT & image & yes & \href{https://github.com/rafaelpadilla/review\_object\_detection\_metrics}{link} \\ \hline

 TracKlinic \cite{Fan_2021_WACV} & 2021 & OT & video & yes & - \\ \hline
Chen et al.
 \cite{chen2021diagnosing} & 2021 & VRD & video & yes & \href{https://github.com/shanshuo/DiagnoseVRD}{link} \\ \hline

 AIDeveloper \cite{krater2021aideveloper} & 2021 & CL & image & yes & \href{https://github.com/maikherbig/AIDeveloper}{link} \\ \hline
 
DETOXER \cite{nourani2022detoxer} & 2022 & CL & video & yes & \href{https://github.com/MahsanNourani/DETOXER}{link} \\ \hline
\end{tabular}
}
\label{tab:tools}
\end{table}
 
\begin{figure}[t]
    \centering  
    \includegraphics[width=.91\linewidth]{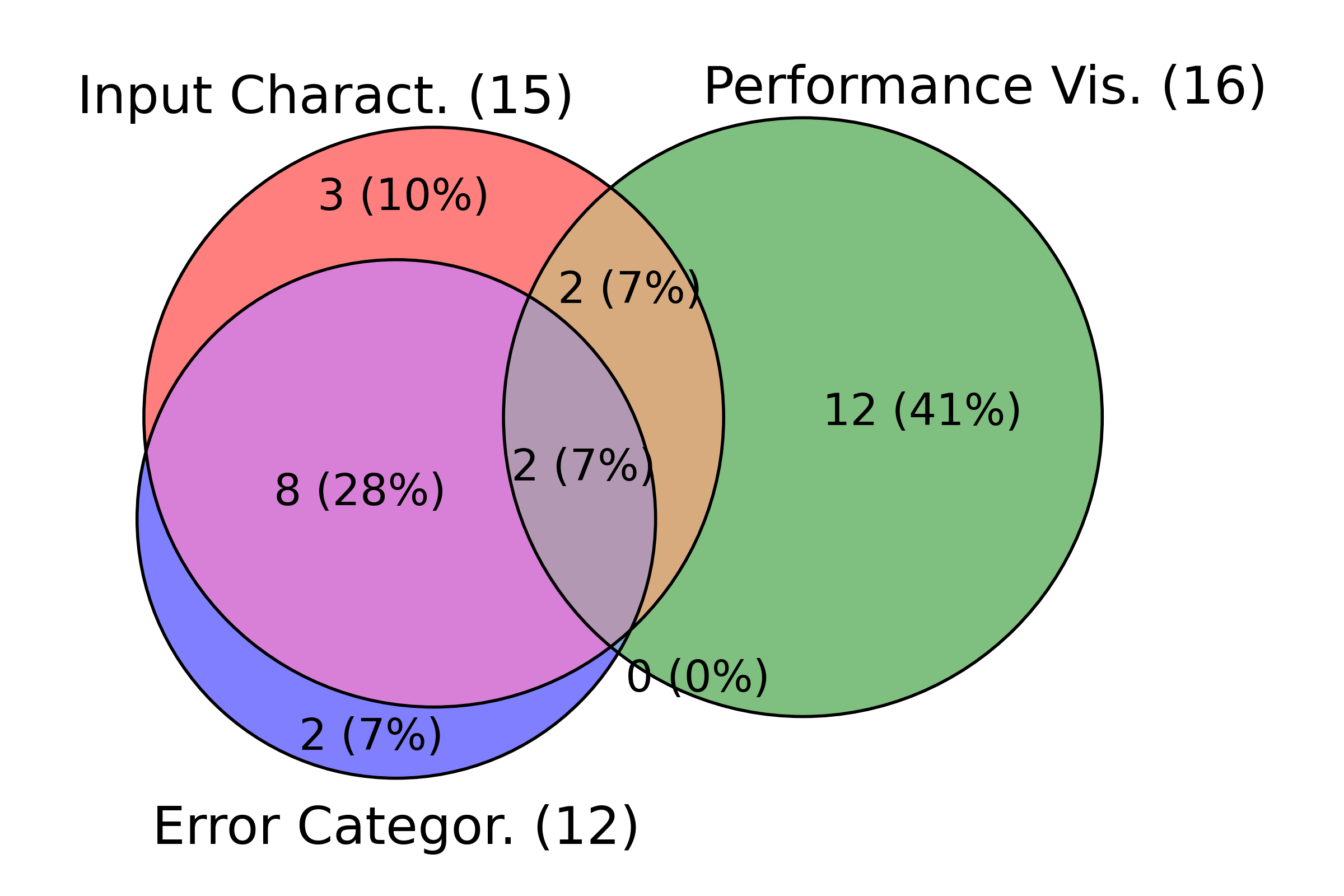}
    \caption{Venn diagram of the distribution of the surveyed tools in three macro-categories: Input characterization oriented, Error categorization oriented and Performance visualization oriented. Each circle represents the tools in one specific macro-category, while the intersections between circles show tools assigned to multiple categories (e.g., the 8 tools in the violet area pertain to both the Input characterization and Error categorization categories)}
    \label{fig:toolscats}
\end{figure}

The tools  can be grouped into three categories, shown in Figure \ref{fig:toolscats},  based on their prevalent approach to  prediction performance diagnosis.  

\begin{figure}[h!]
    \centering
    \includegraphics[width=1\linewidth]{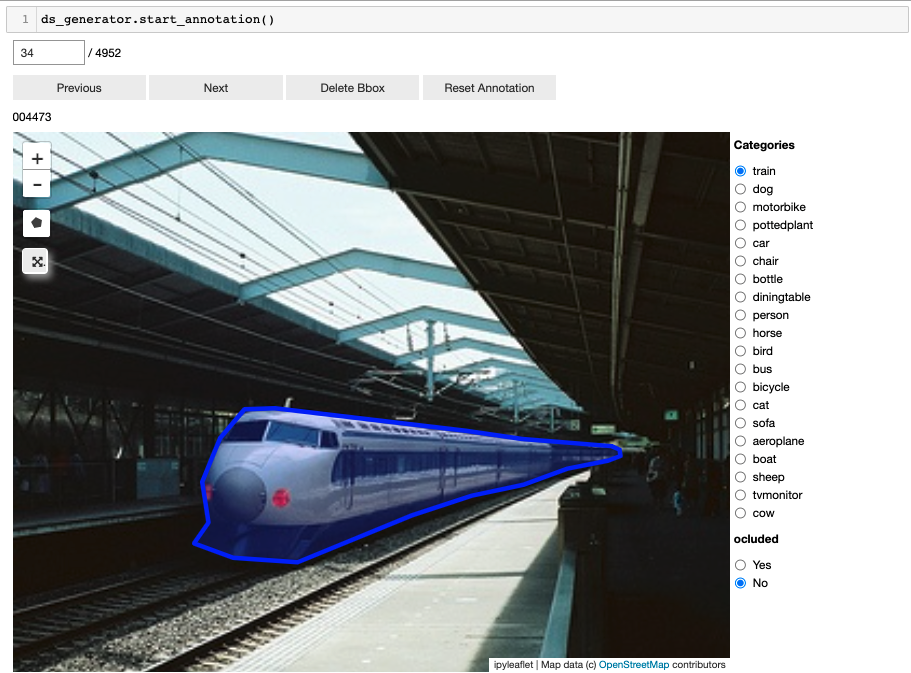}

    \caption{ODIN tool \citep{torres2021odin}: the GUI lets the user annotate the input samples denoting either the class or a custom  property not used for training 
    }
    \label{fig:odin}
\end{figure}

\begin{figure}[h!]
    \centering
   
    \includegraphics[width=1\linewidth]{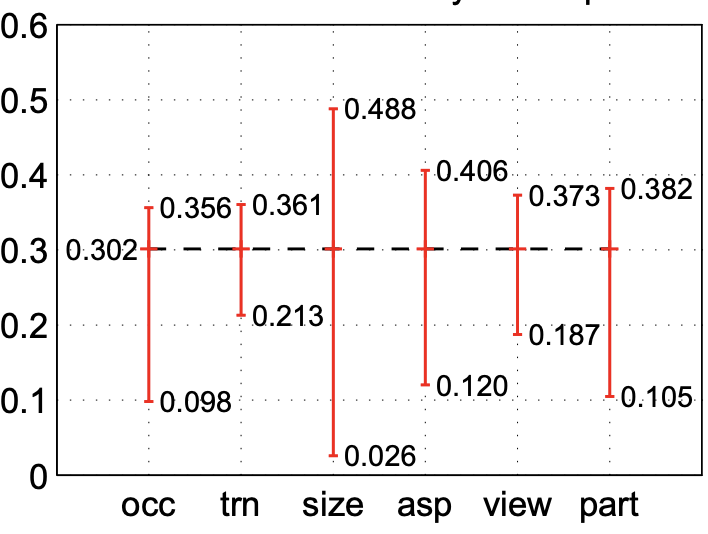}
    
    \includegraphics[width=1\linewidth]{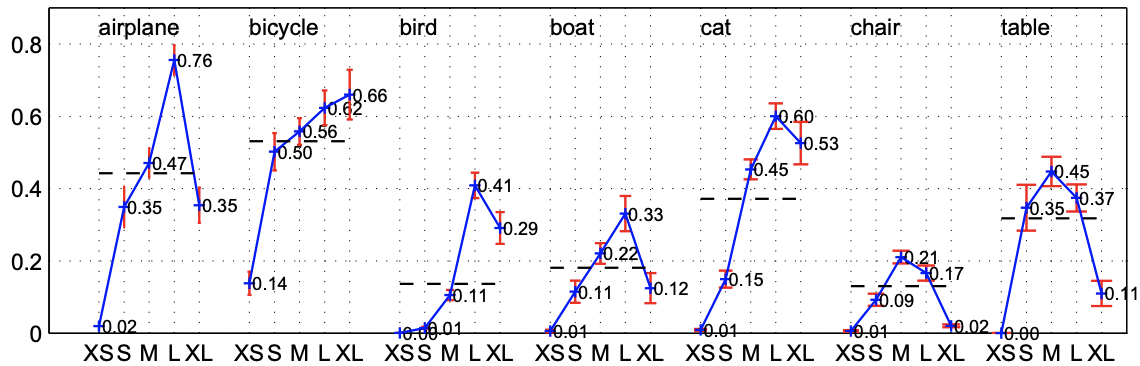}
    
    \caption{
    \citep{hoiem2012diagnosing}  Use of custom properties (occlusion, truncation, bbox size, bbox aspect ratio, view point, part visibility) of the images for performance break down. Property sensitivity and impact analysis diagram (top): it shows the variation of AP for the different properties.  AP break down diagram (bottom): it breaks down the AP metrics by bbox size value
    }
    \label{fig:hoiem}
\end{figure}

\begin{figure}[t]
    \centering
    \includegraphics[width=.5\linewidth]{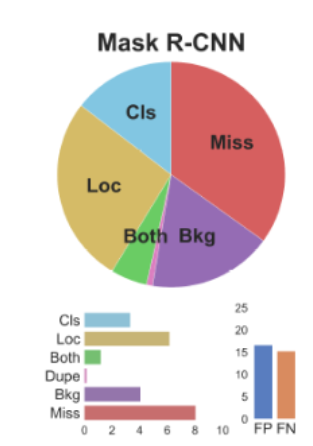}
    \caption{Error categorization in the TIDE  tool \citep{bolya2020tide}: errors are distinguished in the types: categorization, localization, categorization+localization, duplicate detection, background and missed ground truth}
    \label{fig:errorcateg}
\end{figure}

\begin{figure}[htb]
    \centering
    \includegraphics[width=\linewidth]{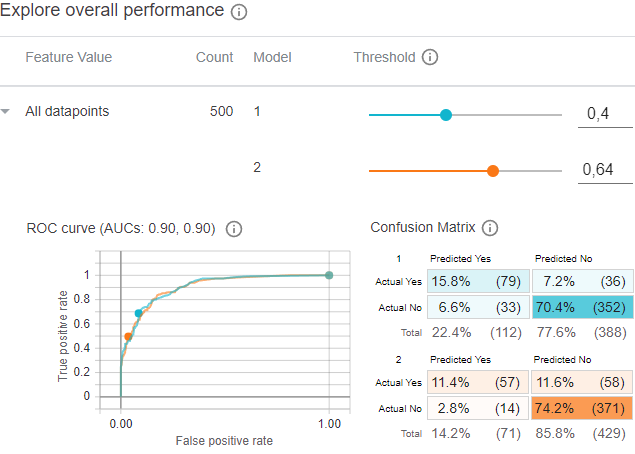}
    \caption{Interactive performance visualization in the What If Tool  \citep{wexler2019if}: the user can interact with the slider to modify the classification threshold  and  view the corresponding point on the curves,  the values of the confusion matrix and the errors}
    \label{fig:whatif}
\end{figure}

\begin{itemize}
    \item \textit{Input characterization oriented}: tools in this class stress the  annotation of the input with custom properties to distinguish which aspects of the input affect the output and cause model failure and performance loss. 
    Figure \ref{fig:odin} shows an example of the interface for annotating images with custom properties and Figure \ref{fig:hoiem} illustrates an example of the use of custom properties  to break down the Average Precision (AP) metrics.
    
    \item \textit{Error categorization  oriented}: tools in this class stress the distinction among different types of errors to quantify  the contribution of an error type to the performance loss. Figure \ref{fig:errorcateg} illustrates an example of error characterization diagram. 
    
    \item \textit{Performance visualization oriented}: tools in this class resort to advanced visualization and interaction  to support the human judgement of the  performance problems. Figure \ref{fig:whatif} shows an example of the  functionality of a performance visualization tool. 
\end{itemize}

\begin{figure}
    \centering
    \includegraphics[width=.95\linewidth]{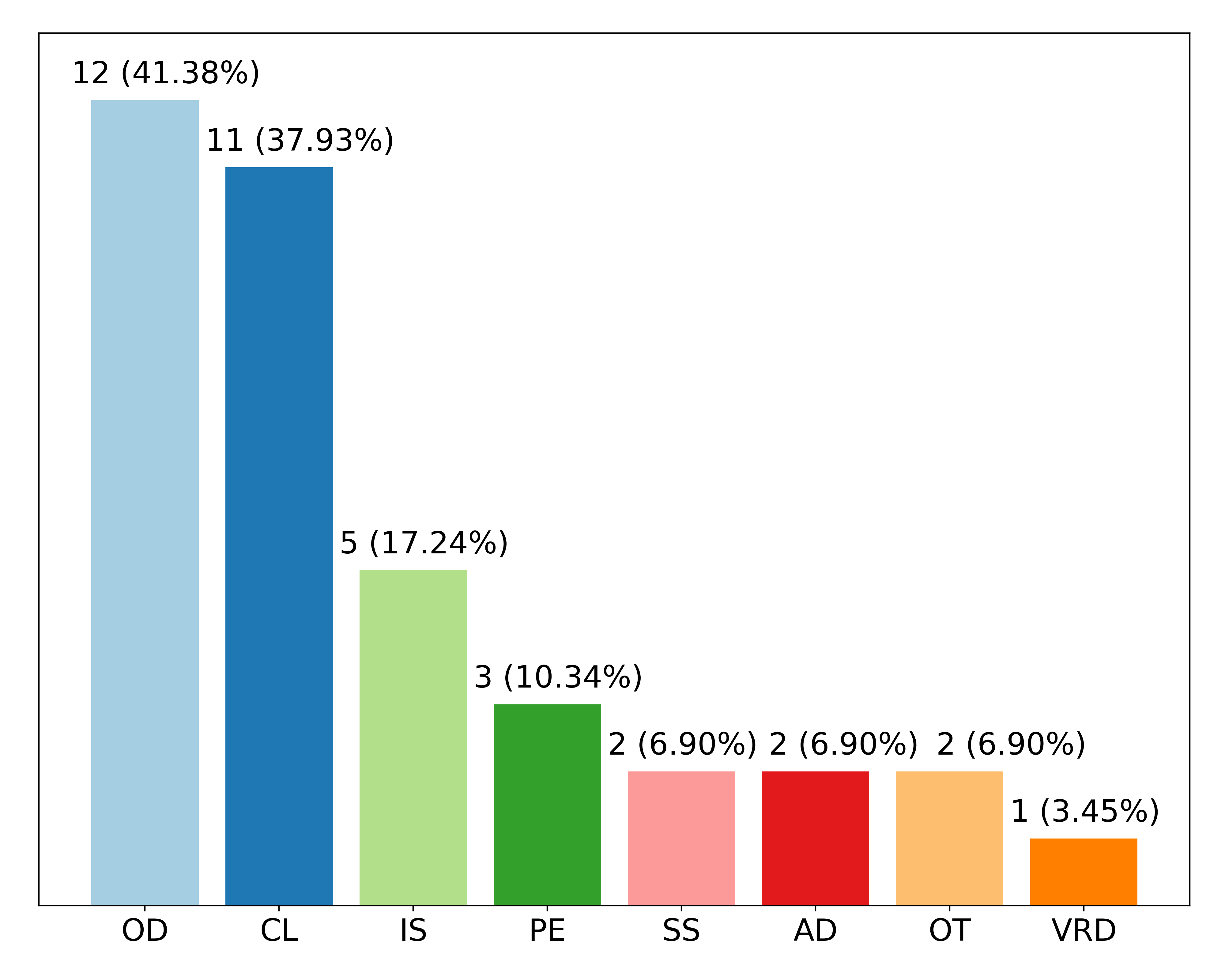}
    \caption{Distribution of the surveyed tools per task}
    \label{fig:toolspertask}
\end{figure}

\begin{figure}
    \centering
    \includegraphics[width=\linewidth]{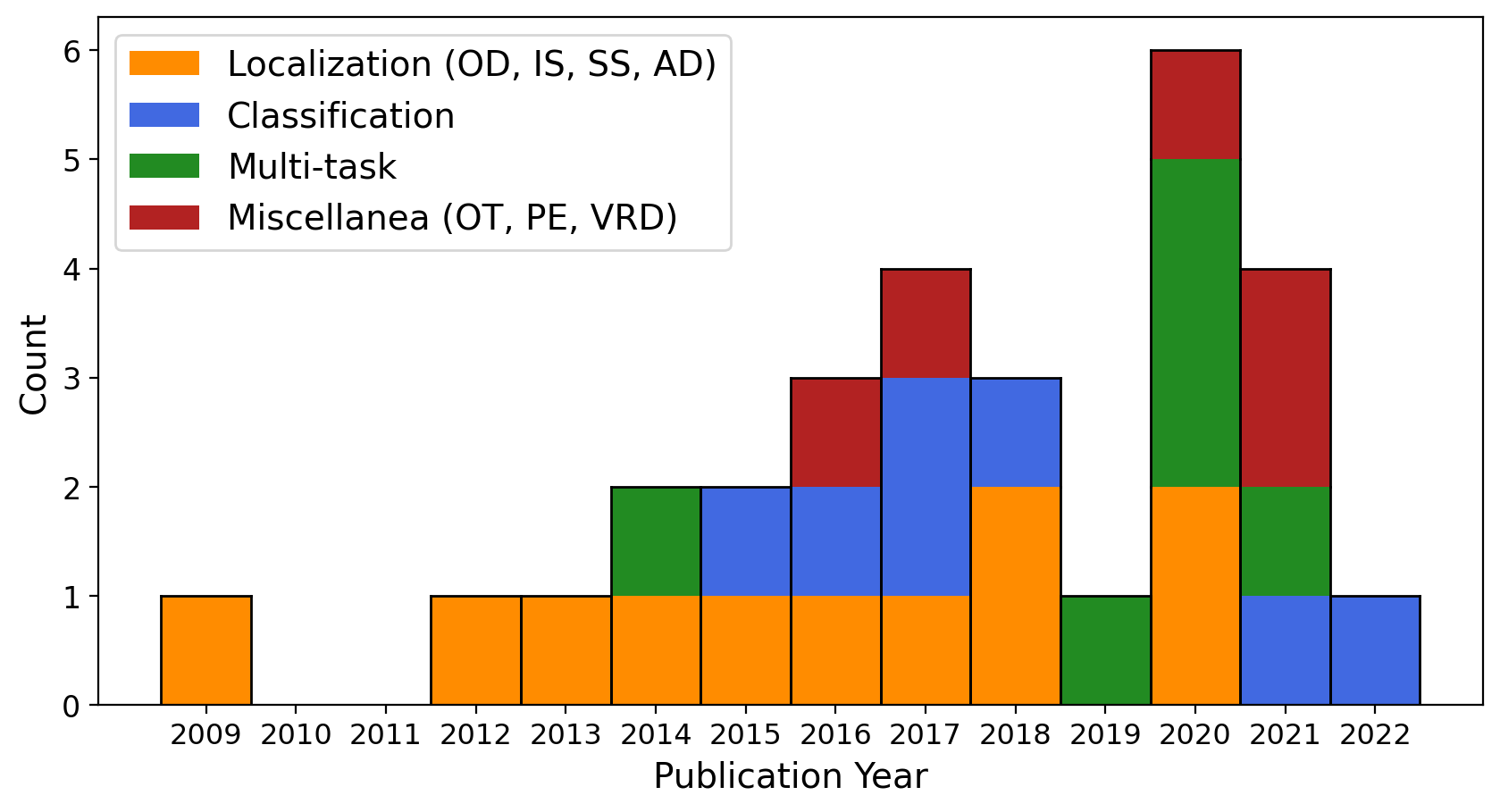}
    \caption{Distribution of the surveyed tools per year and per type of task}
    \label{fig:toolsperyear}
\end{figure}

Figure \ref{fig:toolspertask} shows the distribution of the tools across the CV tasks.  Objection Detection and Classification are the most represented ones. The relevance of OD is not surprising because the pioneering works on black-box  diagnosis    \citep{dollar2009pedestrian} and  \citep{hoiem2012diagnosing} originated from the performance analysis of that task. 
Most tools work with images or with generic inputs, provide open-source code and have been released in the last five years. Figure \ref{fig:toolsperyear} illustrates the distribution over time of the surveyed works. If one does not consider the early 2009 work  \citep{dollar2009pedestrian}, the timeline shows that the interest begun in 2012, the same year in which the research on Deep Learning started its escalation. The idea originated in the CV field for such tasks as object detection and image segmentation and then propagated to other ML applications.  

\subsection{Descriptions of the tools}
In the rest of this section we provide a brief description of the surveyed tools in ascending chronological order.

 \textbf{Dollar et al.:}
The work  \citep{dollar2009pedestrian} presented a data set for pedestrian detection consisting of an annotated video with challenging low resolution images and occluded people. The data set was used to evaluate several detectors and the authors advocated  the use of ad-hoc features of the input to help diagnose errors and support model refinement. To this end, performances were broken down based on specific properties of the input, such as the scale, the aspect ratio of the ground truth bounding boxes and the presence of occlusions of the pedestrians. Although the code was not released, this early work is the first example of a black-box diagnosis tool and inspired the posterior proposals.

\textbf{Hoiem et al.:}
The work in  \citep{hoiem2012diagnosing} pioneered the systematic black-box approach to error analysis in OD tasks and showed the utility of adding extra annotations to  the input besides the labels used for training. The framework exploits a fixed set of diagnosis-oriented object meta-data that can affect the model accuracy, such as: size, parts visibility, aspect ratio,  shape and occlusion (Figure \ref{fig:hoiem}). The authors demonstrate how breaking down standard metrics into sub-metrics linked to a metadata value aids in understanding model faults and in focusing redesign where the margin for improvement is maximal. 

\textbf{Russakovsky et al.:}
The work in  \citep{russakovsky2013detecting} follows the black-box diagnosis method of   \citep{hoiem2012diagnosing} and assesses the performance of several object detectors applied to the ImageNet Large Scale Visual Recognition Challenge (ILSVRC) data set. A first type of analysis breaks down  the performance indicators using  \textit{per-image} properties, such as the number of instances per image and the chance performance of localization (CPL). A second type of analysis shows how the performances are affected by \textit{per-class}  properties, such as the distinctiveness of color and of shape, the instance deformability  and the amount of texture.

\textbf{COCO API:} A step towards the popularization of custom input properties as an aid for error diagnosis is found in the  development framework of the MS COCO data set  \citep{mscoco}. The computation of mean Average Precision (mAP) is differentiated based on the object size: mAP\textsubscript{small}, mAP\textsubscript{medium} and mAP\textsubscript{big}. This distinction  assists the diagnosis of the issues and enables the design of strategies to improve the localization, such as multi-scale object detection  \citep{DENG20183}. In addition to the  the break down of the metrics, the API also allows developers to load any data set that respects the MS COCO format and to visualize both the images and the annotations.

\textbf{Hariharan et al:} The work in  \citep{hariharan2014simultaneous} introduces simultaneous detection and segmentation (SDS) as a novel computer vision task. The authors provide the DNN architecture and a tool for its diagnosis.
Besides the assessment of standard metrics,  error diagnosis is supported by introducing three error classes (localization, confusion with similar classes, and confusion with background) and by computing the impact of each error type on the  performance indicators. 

\textbf{Zhu et.al.:} The authors of  \citep{zhu2015diagnosing} follow the methodology of  \citep{hoiem2012diagnosing} and evaluate object detectors using custom properties. They contrast different methods for creating object proposals on the PASCAL VOC data set. The comparison uses  object  characteristics such as  size, aspect ratio, iconic view, color contrast, shape regularity and texture. The authors also discuss how to exploit objects properties to investigate model limitations and show the sensitivity of the model to the  characteristics of the objects.

\textbf{ModelTracker:} The work in  \citep{amershi2015modeltracker} investigates the performances of a classifier with a black-box approach  that combines  metrics summaries and  interactive visualizations. Binary predictions are color-coded and arranged by classification score. The analysis of results is facilitated by tagging input samples with custom properties and by highlighting samples similarity and outliers.

\textbf{Redondo et al.:} The authors of  \citep{redondo2016pose} propose a diagnostic tool tailored to the study of pose estimation errors. The tool examines the effects of custom properties (aspect ratio, size, visibility of parts) on the detection and pose estimation performances and highlights the impact of different types of pose-related False Positives. The authors analyze four state-of-the-art object detection and posture estimation models to uncover flaws and recommend improvements.

\textbf{Prospector:} The work
in  \citep{krause2016interacting} describes a web-based tool that implements counterfactual analysis. The tool implements a partial dependence technique for determining the impact of each input feature on the DNN results. The developer can apply changes to the input data and measure the impact on the output. The system  suggests the shift in the value of each input feature that would lead to the greatest performance improvement. The diagnostic utility of the approach is demonstrated in a diabetes prediction task.

\textbf{Zhang et al.:}
In  \citep{zhang2016far} the authors apply error diagnosis to the state-of-the-art pedestrian detection algorithms. They enhance the annotations of the Caltech  \citep{dollar2009pedestrian} data set and study both False Positives (FPs) and False Negatives (FNs) with different error categories. FP errors are distinguished into localization, background, and annotation.  FN errors are categorized into scale, viewpoint, occlusion, and other types. The authors also analyze the impact of FPs on performances. 

\textbf{Ronchi et al.:}
The work in  \citep{ruggero2017benchmarking} applies the approach of  \citep{hoiem2012diagnosing} to the Multi-Instance Pose Estimation task. Three error types are defined (localization, scoring and background) and the impact of three challenging factors is studied (occlusion, crowding and size). Their tool visualizes the distribution of errors for each key point and highlights the improvement in the Precision-Recall curve obtainable by correcting specific types of errors.

\textbf{Explanation Explorer:}
In  \citep{krause2017workflow}, the authors of Prospector  \citep{krause2016interacting}  describe a novel tool for the assessment and interpretability of binary classifiers. Their approach comprises three steps: 1) the display of classical performance metrics and analyses; 2) the explanation generation, which computes the features of the input samples that impact the outcome most significantly; 3) the interactive visualization of the explanations. The visualization is organized in three stages: 1) outcome-level, focusing on the overall accuracy; 2) feature-level, presenting the explanations along with the corresponding features; and 3) instance-level, which allows the user to analyze each instance and derive hypotheses about the classifier failures. The authors advocate that visual analytics should play a major role in error diagnosis but also highlight that not all failures can be rectified by training a stronger model because some errors require bias mitigation in the original data set.

\textbf{Squares:}
Squares  \citep{ren2016squares} is a tool for the interactive performance analysis of multi-class single-label classifiers. The classes and the corresponding instances are displayed on the same row with a distinctive color. The observations are ordered by their prediction confidence score and grouped. The first group represents the FNs whereas the second cluster comprises both the True Positives (TPs) (highlighted with the color of their class) and the FPs (highlighted with the color of their true class). When an observation is selected from one class, its representations in the other classes are emphasized visually, thus allowing a comparison among the different predictions of the same sample.

\textbf{Sigurdsson et al.:}
The work in  \citep{sigurdsson2017actions} surveys the state-of-the-art in the action detection task and compares the existing methods.
The evaluation is based on the categorization of errors in four classes: boundary, other class with the same object, other class with the same verb, other class with neither and no class. A further analysis evaluates the models w.r.t. the complexity of the objects/verbs that characterize a category. Finally, the authors examine the impact of two specific features of the input: the temporal extent of the action and the presence of people.

\textbf{DETAD:}
The focus of  \citep{alwassel2018diagnosing} is on the identification of temporal actions in videos. The diagnosis tool enables the analysis of FPs and FNs and the estimation of the sensitivity of mAP-based metrics to six action characteristics: length, context distance, agreement, coverage, context size and number of instances.

\textbf{Nekrasov et al.:}
In  \citep{nekrasov2018diagnostics} the authors apply the black-box diagnostic strategy of  \citep{hoiem2012diagnosing} to the semantic segmentation task.

\textbf{Manifold:}
The authors of  \citep{zhang2018manifold} discuss an interactive framework for the evaluation and debugging of ML models. An agreement analysis function enables the comparison of model pairs by highlighting the similarities and differences of their predictions. A feature distribution function permits the selection of a subset of the samples and  measures the intra-group similarity based on the occurrence frequency of each feature.

\textbf{What If Tool:} As the name suggests, the tool described in
 \citep{wexler2019if} allows researchers to analyze the performances of ML systems in hypothetical situations by visualizing the effect of several features on different models and on different subsets of the input data. Among all the surveyed tools, this is the only one providing also a fairness analysis that highlights bias in the input data set.  Other functionalities support data point editing, counterfactual reasoning, performance measures for  classification  and regression (Figure \ref{fig:whatif}) and data set fairness optimization.

\begin{figure*}[t]
    \centering
    \includegraphics[width=.8\linewidth]{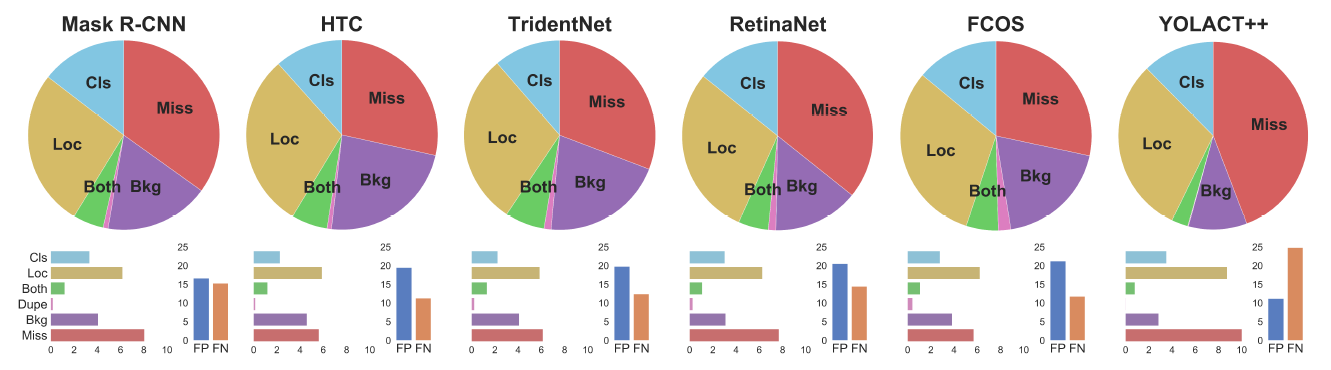}
    \caption{Error categorization and model comparison in TIDE  \citep{bolya2020tide}}
    \label{fig:tide}
\end{figure*}

\textbf{TIDE:}
The tool illustrated in  \citep{bolya2020tide} supports error diagnosis in object detection and instance segmentation. The tool is applicable to multiple data sets and output formats. Similarly to  \citep{hoiem2012diagnosing} it categorizes  detection and segmentation errors  and provides compact error summaries and impact reports. For example, it includes the direct comparison of results by different models  (Figure \ref{fig:tide}). Unlike other tools such as  \citep{hoiem2012diagnosing,torres2020odin}, TIDE purposely avoids resorting to properties of the input other than those used for training. 

\textbf{ODIN:}
The \framework framework  \citep{torres2020odin, torres2021odin} aims at generalizing and integrating into a unique solution the previous approaches to DNN black-box diagnosis for classification, object detection and instance segmentation. \framework allows the addition of custom properties to the input, supports the plug-in of user-defined performance indicators and implements a wide range of metrics and analysis reports off-the-shelf (Figure \ref{fig:odin}). It can be used to study both model performance and data set bias. It combines error impact sensitivity and confidence calibration. The latter analysis evaluates the similarity of the distribution of predictions to the real probability distribution of the input data. The tool also comprises a Graphical User Interface (GUI) for annotating the input with custom properties.

\textbf{Padilla et al.:}
The authors of  \citep{padilla2020survey} discuss the most commonly used metrics for object detection and provide the implementation code in a toolkit. The work in  \citep{padilla2021comparative} describes the most recent release, which makes the toolkit independent of the input formats, adds more bounding box formats, and includes novel spatio-temporal metrics for object detection in video.

\textbf{TF-GraF:} The goal of the work in 
 \citep{yoon2020tensorflow} is to create an easy-to-use Tensorflow object detection environment, simplifying the installation and set-up and the coding and execution of the workflows. The tool supports pre-processing, training and evaluation with the MS COCO metrics. It incorporates the best known object detection and instance segmentation architectures (Faster RCNN, SSD, Mask RCNN) and provides the visualization of the training and test data sets. Non-experts can configure, train, and assess DNNs with no programming.

\textbf{Boxer:}
The work in  \citep{gleicher2020boxer} presents a tool called Boxer for comparing the performances of different classifiers. The system supports the selection of metrics, the grouping of training and test data into \quoted{boxes} based on selected features and the comparative visualization of the outputs. To evaluate data quality and bias, a novel method based on set algebra is presented to link views and analyze multiple data subsets. The authors demonstrate the tool utility in a variety of use cases and discuss strategies for dealing with very large data sets.

\textbf{OpenVINO DL Workbench:} 
The work in  \citep{demidovskij2021openvino} focuses on model training, analysis, optimization, evaluation and deployment, covering the entire model development workflow with a hybrid black-box and white-box approach. The workbench includes an Accuracy Checker that implements a black-box analysis for classification, regression, and object detection tasks by computing the most common metrics for the whole data set and per class. It also supports the open-box analysis, for example to evaluate and improve model performance in terms of execution time and memory consumption. 

\textbf{TracKlinic:} The work in
 \citep{Fan_2021_WACV} studies the factors that challenge object tracking in videos. Custom properties can be manually associated with the video frames to specify seven common error-inducing factors: occlusion, rotation, out-of-view, background clutter, illumination variation, shape variation, and motion blur. The proposed diagnosis tool exploits the Intersection over Union (IoU) base metrics to analyze the failure rates and the success scores of ten state-of-the-art architectures applied to three benchmark data sets manually annotated with the above mentioned factors. Per-frame proposals of alternative models can be visualized together and compared with the ground truth annotations. The diagnosis results show that most models fail when complex situations occur, such as out-of-view transitions and shape variations.

\textbf{Chen et al.:}
The work in  \citep{chen2021diagnosing} focuses on the relation detection task in videos and assesses the state-of-the-art detectors over two benchmarks (ImageNet-VidVRD  \citep{shang2017video} and VidOR  \citep{shang2019annotating}). The authors categorize the FPs  and compute their impact over the Average Precision, analyze the FN distribution across different input characteristics (e.g., the video length, the number of subject/predicate/object instances, the subject/object pixel scale) and compute the performance gain achievable by removing each error type.

\textbf{AIDeveloper:} The work in \citep{krater2021aideveloper} presents an open-source software supporting the entire development process of an image classification model: dataset definition and visualization, model training and optimization  and performance evaluation. The tool includes an easy-to-use GUI that allows researchers to develop  DL-powered applications without coding.

\textbf{DETOXER:} The work in \cite{nourani2022detoxer} is an interactive visual analytics tool for debugging Temporal Multi-Label Classification models in multiple videos. The authors designed it in order to provide three different levels of granularity for explanations and evaluation: frame-level analysis offering a compact visualization of all the categories; video-level explanation providing an overview of the errors (false positives and false negatives) for each video; and, global-level summary revealing error trends across all the analyzed videos.

 \section{Comparison of diagnostic tools}\label{sec:tool-cmp}

For each tool described in Section \ref{sec:tool-desc} the supported metrics and types of analysis were extracted, resulting in more than 70 options.
For ease of comparison  four tables  are introduced, one  for each family  of homogeneous metrics/analyses: generic multi-task (Table \ref{tab:comp_generic}), classification (Table \ref{tab:comp_cl}), localization (Table \ref{tab:comp_loc}), and a miscellaneous category grouping the functions found in the less frequent  tools for object tracking and pose estimation (Table \ref{tab:comp_ot_pe}). The rows specify the metrics/analysis and the columns the tools that support them, sorted in chronological order. For space reasons, the tool  or authors' names are omitted  but they can be recovered from  the corresponding reference. Each cell specifies if the  option is offered by the specified tool. The symbol \quoted{-} means that the row is not relevant for the specific tool. For example, a metrics  specific for pose estimation is not relevant  for tools focused on object tracking. Note that when some base metrics is used to compute a derived metrics (e.g., IoU and AP in Table \ref{tab:comp_loc})  the  row of the base metrics contains the \checkmark value only when the tool exposes the base metrics explicitly.

\subsection{Multi-task metrics and analyses}
Table \ref{tab:comp_generic} lists 16 general metrics that apply to all the considered tasks and the 23 tools that implement them. Only Accuracy is well represented in the surveyed classification tools.   Other standard ML metrics (e.g., precision, recall, F1-score and AUCs) are present only in 4 or 5 tools, less  than one may expect. A similar consideration  also holds for the  Precision-Recall, F1 and ROC curves,  usually employed  to visualize and compare performances and optimize the hyperparameters. Only 7 out of 23 tools offer such features. The most represented types of analysis are those related to FPs and FNs, which are the most common starting points for error diagnosis.

\begin{table*}[htb]
\centering
\normalsize
\caption{Multi-task metrics and analyses.}
\label{tab:comp_generic}
\resizebox{\textwidth}{!}{
\begin{tabular}[c]{ p{2cm} K{0.55cm} K{0.55cm} K{0.55cm} K{0.55cm} K{0.55cm} K{0.55cm} K{0.55cm} K{0.55cm} K{0.55cm} K{0.55cm} K{0.55cm} K{0.55cm} K{0.55cm} K{0.55cm} K{0.55cm} K{0.55cm} K{0.55cm} K{0.55cm} K{0.55cm} K{0.55cm} K{0.55cm} K{0.55cm} K{0.55cm}}
\hline
\textbf{Metric / Analysis} & \textbf{\cite{hoiem2012diagnosing}} & \textbf{\cite{russakovsky2013detecting}} & \textbf{\cite{mscoco}} & \textbf{\cite{hariharan2014simultaneous}} & \textbf{\cite{zhu2015diagnosing}} & \textbf{\cite{redondo2016pose}} & \textbf{\cite{krause2016interacting}} & \textbf{\cite{zhang2016far}} & \textbf{\cite{ruggero2017benchmarking}} & \textbf{\cite{krause2017workflow}} & \textbf{\cite{sigurdsson2017actions}} & \textbf{\cite{alwassel2018diagnosing}} & \textbf{\cite{nekrasov2018diagnostics}} & \textbf{\cite{wexler2019if}} & \textbf{\cite{bolya2020tide} } & \textbf{\cite{torres2020odin}} & \textbf{\cite{padilla2020survey}} & \textbf{\cite{gleicher2020boxer}} & \textbf{\cite{demidovskij2021openvino}} & \textbf{\cite{Fan_2021_WACV}} & \textbf{\cite{chen2021diagnosing}} & \textbf{\citep{krater2021aideveloper}} &
\textbf{\citep{nourani2022detoxer}}\\ \hline 
Accuracy & - & \textbf{\checkmark} & - & - & - & - & \textbf{\checkmark} & - & - & \textbf{\checkmark} & - & - & \textbf{\checkmark} & \textbf{\checkmark} & - & \textbf{\checkmark} & - & \textbf{\checkmark} & \textbf{\checkmark} & - & - & \textbf{\checkmark} & \textbf{\checkmark}\\ \hline 
Precision & & & & & & & & & & & & & & & & \textbf{\checkmark} & & \textbf{\checkmark} & \textbf{\checkmark} & - & & \textbf{\checkmark} & \\ \hline 
Recall & & & & & \textbf{\checkmark} & & & & & & & & & & & \textbf{\checkmark} & & \textbf{\checkmark} & \textbf{\checkmark} &  - & & \textbf{\checkmark} & \\ \hline 
ROC Curve & - & - & - & - & - & - & & - & - & \textbf{\checkmark} & - & - & & \textbf{\checkmark} & - & \textbf{\checkmark} & - & & & - & - & \textbf{\checkmark} &\\ \hline 
ROC AUC & - & - & - & - & - & - & & - & - & \textbf{\checkmark} & - & - & & \textbf{\checkmark} & - & \textbf{\checkmark} & - & & \textbf{\checkmark} &  - & -  & \textbf{\checkmark} & \\ \hline 
PR curve & & & & \textbf{\checkmark} & & & & & \textbf{\checkmark} & & & & & \textbf{\checkmark} & & \textbf{\checkmark} & \textbf{\checkmark} & & &  - & & \textbf{\checkmark} & \\ \hline 
PR AUC & & & & & & & \textbf{\checkmark} & & & & & & & \textbf{\checkmark} & & \textbf{\checkmark} & & & &  - & & &\\ \hline 
F1 Score & & & & & & & & & & & & & & & & \textbf{\checkmark} & & \textbf{\checkmark} & \textbf{\checkmark} &  - & & \textbf{\checkmark} &\\ \hline 
F1 Curve & & & & & & & & & & & & & & & & \textbf{\checkmark} & & & &  - & & &\\ \hline 
F1 AUC & & & & & & & & & & & & & & & & \textbf{\checkmark} & & & &  - & & &\\ \hline 
\# FP & & & & & & & & & & & & & & \textbf{\checkmark} & & \textbf{\checkmark} & & & &  & & & \textbf{\checkmark}\\ \hline 
\# FN & & & & & & & & & & & & & & \textbf{\checkmark} & & \textbf{\checkmark} & & & &  & & & \textbf{\checkmark}\\ \hline 
FN Analysis & & & \textbf{\checkmark} & & & & & \textbf{\checkmark} & \textbf{\checkmark} & & & \textbf{\checkmark} & & & \textbf{\checkmark} & \textbf{\checkmark} & & & &  & \textbf{\checkmark} & &\\ \hline 
FP Analysis & \textbf{\checkmark} & & \textbf{\checkmark} & \textbf{\checkmark} & & \textbf{\checkmark} & & \textbf{\checkmark} & \textbf{\checkmark} & & \textbf{\checkmark} & \textbf{\checkmark} & \textbf{\checkmark} & & \textbf{\checkmark} & \textbf{\checkmark} & & & & & \textbf{\checkmark} & &\\ \hline 
TP Analysis & & & & & & & & & & & & & & & & \textbf{\checkmark} & & & &  & & &\\ \hline 
Reliability Analysis & & & & & & & & & & & & & & & & \textbf{\checkmark} & & & &  \textbf{\checkmark} & & &\\ \hline 
\end{tabular}
}
\end{table*} 
\subsection{Classification metrics and analyses}
Table \ref{tab:comp_cl} lists 8 classification-specific metrics implemented in 8 frameworks. The Confusion Matrix is implemented by 7 out of 8 tools, whereas Error Rate, Mean Error (ME), Mean Absolute Error (MAE), Mean Squared Error (MSE), Odds Ratio and the True Negative (TN) analysis are implemented by only one proposal.

\begin{table*}[htb]
\centering
\footnotesize
\caption{Classification-specific metrics and analyses.}
\label{tab:comp_cl}
\begin{tabular}[c]{ l c c c c c c c c}
\hline
\textbf{Metric / Analysis}  & 
\textbf{\cite{krause2017workflow}} &
\textbf{\cite{ren2016squares}}  & 
\textbf{\cite{zhang2018manifold}} &
\textbf{\cite{wexler2019if}}  & 

\textbf{\cite{torres2020odin}}  & 
\textbf{\cite{gleicher2020boxer}}  & \textbf{\cite{demidovskij2021openvino}} & \textbf{\citep{krater2021aideveloper}}\\ \hline

Error Rate & & &  &  & \textbf{\checkmark} & & &\\ \hline

Confusion Matrix & \textbf{\checkmark} & \textbf{\checkmark} & \textbf{\checkmark} & \textbf{\checkmark}  & \textbf{\checkmark} & \textbf{\checkmark} & & \textbf{\checkmark}\\ \hline

Mean Error & & & & \textbf{\checkmark} & & & & \\ \hline

Mean Absolute Error & & & & \textbf{\checkmark} & & & &\\ \hline

Mean Squared Error & & & & \textbf{\checkmark} & & & &\\ \hline

Odds Ratio & \textbf{\checkmark} & & & & & & &\\ \hline

Matthews Correlation Coefficient & & & &  & & \textbf{\checkmark} & \textbf{\checkmark} &\\ \hline

TN Analysis & & & & & & \textbf{\checkmark} & &\\ \hline

\end{tabular}
\end{table*}
 
\subsection{Localization metrics and analyses}
Table \ref{tab:comp_loc} presents the 6 localization-specific metrics and the 19 tools that implement them. The surveyed frameworks support a variety of localization tasks:  OD, IS, SS, AD and PE. The review shows that  there is little consensus among localization-oriented frameworks about which metrics are essential and should be provided off-the-shelf. The implementation by tools concentrates only on the  metrics commonly required by the most popular CV benchmarks: Average Precision (IoU) for OD and IS;  other useful metrics, both general and localization-specific, such as Miss Rate, Average Recall (IoU) or F1 Score are implemented rather infrequently by the object localization tools.  
\begin{table*}[htb]
\centering
\small
\caption{Localization-specific metrics and analyses.}
\label{tab:comp_loc}
\resizebox{\textwidth}{!}{
\begin{tabular}[c]{ l K{0.55cm} K{0.55cm} K{0.55cm} K{0.55cm} K{0.55cm} K{0.55cm} K{0.55cm} K{0.55cm} K{0.55cm} K{0.55cm} K{0.55cm} K{0.55cm} K{0.55cm} K{0.55cm} K{0.55cm} K{0.55cm} K{0.55cm} K{0.55cm} K{0.55cm} K{0.55cm}  }
\hline
\textbf{Metric / Analysis} & \textbf{\cite{dollar2009pedestrian}} & \textbf{\cite{hoiem2012diagnosing} } & \textbf{\cite{mscoco}} & \textbf{\cite{hariharan2014simultaneous}} & \textbf{\cite{zhu2015diagnosing} } & \textbf{\cite{redondo2016pose}} & \textbf{\cite{zhang2016far}} & \textbf{\cite{ruggero2017benchmarking}} & \textbf{\cite{sigurdsson2017actions}} & \textbf{\cite{alwassel2018diagnosing}} & \textbf{\cite{nekrasov2018diagnostics} } & \textbf{\cite{bolya2020tide} }  & \textbf{\cite{torres2020odin}} & \textbf{\cite{padilla2020survey}} & \textbf{\cite{yoon2020tensorflow}}& \textbf{\cite{demidovskij2021openvino}} & \textbf{\cite{padilla2021comparative}} & \textbf{\cite{Fan_2021_WACV}} & \textbf{\cite{chen2021diagnosing}} \\ \hline
Mean IoU & & & & \textbf{\checkmark} & & - & - & & & & \textbf{\checkmark} & &  & & &\textbf{\checkmark} & & & \\ \hline
Average Precision (IoU) & & \textbf{\checkmark} & \textbf{\checkmark} & & \textbf{\checkmark} & \textbf{\checkmark} & \textbf{\checkmark} & & \textbf{\checkmark} & \textbf{\checkmark} & & \textbf{\checkmark} & \textbf{\checkmark} & \textbf{\checkmark} & \textbf{\checkmark} & \textbf{\checkmark} & \textbf{\checkmark} & & \textbf{\checkmark} \\ \hline
Average Recall (IoU) & & & \textbf{\checkmark} & & & & & & & & &  & & & \textbf{\checkmark} & \textbf{\checkmark} & \textbf{\checkmark} & & \\ \hline
Miss Rate & \textbf{\checkmark} & & & & & & & \textbf{\checkmark} & - & - & & &  & & & \textbf{\checkmark} & & & \\ \hline
Localization Latency & & & & \textbf{\checkmark} & & & & & & & & & & & & & & & \\ \hline
IoU Analysis & & & & \textbf{\checkmark} & \textbf{\checkmark} & \textbf{\checkmark} & & & & & & \textbf{\checkmark} &   \textbf{\checkmark} & & & & 
\textbf{\checkmark} & & \\ \hline
\end{tabular}
}
\end{table*} 
\subsection{Object Tracking and Pose Estimation metrics and analyses}
Table \ref{tab:comp_ot_pe} lists task-specific metrics found in tools designed for OT and PE. Also in this case a lack of consensus about a common set of  relevant metrics can be observed. For example both   \citep{ruggero2017benchmarking} and  \citep{redondo2016pose} focus on PE but do not share any of the task-specific metrics.
\begin{table*}[htb]
\centering
\footnotesize
\caption{Object Tracking and Pose Estimation metrics and analyses.}
\label{tab:comp_ot_pe}
\begin{tabular}{ l c c c c c }
\hline
\textbf{Metric / Analysis}  & \textbf{\cite{redondo2016pose} }  & \textbf{\cite{ruggero2017benchmarking}}    & \textbf{\cite{padilla2021comparative}}  & \textbf{\cite{Fan_2021_WACV}}  \\\hline
Success Score (OT) & - & - & & \textbf{\checkmark} \\ \hline
Failure Rate (OT) & - & - & & \textbf{\checkmark} \\ \hline
Consistency Analysis (OT) & - & - & & \textbf{\checkmark} \\ \hline
Spatio-Temporal Tube Average Precision (STT-AP) (OT) & - & - &  \textbf{\checkmark} & \\ \hline
Pose Estimation Average Precision (PE) & \textbf{\checkmark} & &  - & - \\ \hline
Average Viewpoint Precision(PE) & \textbf{\checkmark} & &  - & - \\ \hline
Average Orientation Similarity(PE) & \textbf{\checkmark} & &  - & - \\ \hline
Mean Angle Error (PE) & \textbf{\checkmark} & &  - & - \\ \hline
Median Angle Error (PE) & \textbf{\checkmark} & &  - & - \\ \hline
Object Keypoint Similarity (PE) & & \textbf{\checkmark} &  - & - \\ \hline
\end{tabular}
\end{table*} 
\subsection{Error categorization metrics and analyses}
Table \ref{tab:comp_err} lists 12 tools that implement error categorization and define 14 types of errors. Error categorization is provided only by OD, IS or PE tools and is not yet common for other tasks. As discussed in the surveyed works, error categorization unveils specific factors associated with model failure that are difficult to extract from aggregated metrics alone. The error categorization tools exploit this feature for impact analysis and highlight the performance gain obtainable by removing or mitigating a specific type of error. 

\begin{table*}[htb]
\centering
\footnotesize
\caption{Error categorization metrics and analysis.}
\label{tab:comp_err}
\resizebox{\textwidth}{!}{
\begin{tabular}{ l c c c c c c c c c c c c }
\hline
\textbf{Metric / Analysis} & \textbf{\cite{hoiem2012diagnosing}} & \textbf{\cite{hariharan2014simultaneous}} & \textbf{\cite{redondo2016pose}} & \textbf{\cite{zhang2016far}} & \textbf{\cite{ruggero2017benchmarking}} & \textbf{\cite{sigurdsson2017actions}} & \textbf{\cite{alwassel2018diagnosing}} & \textbf{\cite{nekrasov2018diagnostics}} & \textbf{\cite{bolya2020tide}} & \textbf{\cite{torres2020odin}} & \textbf{\cite{Fan_2021_WACV}} & \textbf{\cite{chen2021diagnosing}} \\ \hline
Errors Categorization (ET) & \textbf{\checkmark} & \textbf{\checkmark} & \textbf{\checkmark} & \textbf{\checkmark} & \textbf{\checkmark} & \textbf{\checkmark} & \textbf{\checkmark} & \textbf{\checkmark} & \textbf{\checkmark} & \textbf{\checkmark} & \textbf{\checkmark} & \textbf{\checkmark} \\ \hline
ET: Classification & \textbf{\checkmark} & & & & & & & & \textbf{\checkmark} & & & \textbf{\checkmark} \\ \hline
ET: Localization & \textbf{\checkmark} & \textbf{\checkmark} & & \textbf{\checkmark} & \textbf{\checkmark} & & \textbf{\checkmark} & \textbf{\checkmark} & \textbf{\checkmark} & \textbf{\checkmark} & & \textbf{\checkmark} \\ \hline
ET: Classification + Localization & & & & & & & & & \textbf{\checkmark} & & & \\ \hline
ET: Duplicated & \textbf{\checkmark} & & & & & & \textbf{\checkmark} & & \textbf{\checkmark} & & & \textbf{\checkmark} \\ \hline
ET: Missed Ground Truth & & \textbf{\checkmark} & & & & \textbf{\checkmark} & & & \textbf{\checkmark} & & & \textbf{\checkmark} \\ \hline
ET: Confusion with background & \textbf{\checkmark} & \textbf{\checkmark} & - & \textbf{\checkmark} & \textbf{\checkmark} & & \textbf{\checkmark} & \textbf{\checkmark} & \textbf{\checkmark} & \textbf{\checkmark} & \textbf{\checkmark} & \textbf{\checkmark} \\ \hline
ET: Confusion with similar class & \textbf{\checkmark} & \textbf{\checkmark} & - & & & & & \textbf{\checkmark} & & \textbf{\checkmark} & & \\ \hline
ET: Confusion withn similar class & \textbf{\checkmark} & & - & & & & & \textbf{\checkmark} & & \textbf{\checkmark} & & \\ \hline
ET: Other & \textbf{\checkmark} & \textbf{\checkmark} & \textbf{\checkmark} & \textbf{\checkmark} & \textbf{\checkmark} & \textbf{\checkmark} & & & & \textbf{\checkmark} & \textbf{\checkmark} & \textbf{\checkmark} \\ \hline
ET: Opposite (PE) & - & & \textbf{\checkmark} & - & & - & - & - & - & - & - & - \\ \hline
ET: Nearby (PE) & - & & \textbf{\checkmark} & - & & - & - & - & - & - & - & - \\ \hline
ET: Confusion with other class with same object & - & - & - & - & - & \textbf{\checkmark} & & - & - & - & - & - \\ \hline
ET: Confusion with other class with same verb & - & - & - & - & - & \textbf{\checkmark} & & - & - & - & - & - \\ \hline
ET: Boundary & - & - & - & - & - & \textbf{\checkmark} & & - & - & - & - & - \\ \hline
Error Contribution Analysis & \textbf{\checkmark} & \textbf{\checkmark} & \textbf{\checkmark} & \textbf{\checkmark} & \textbf{\checkmark} & & \textbf{\checkmark} & \textbf{\checkmark} & \textbf{\checkmark} & \textbf{\checkmark} & & \textbf{\checkmark} \\ \hline
\end{tabular}
}
\end{table*} 
\subsection{Additional features}
Table \ref{tab:comp_feat} presents 14   features not related to any specific metrics. Some functions let the user inspect the predictions at different levels of granularity and display the distribution of classes and properties in the input data set. Other functions refer to the availability and extensibility of custom properties and metrics. A last group contains functions for model comparison and data set annotation and visualization. 
The multi-level analysis and the comparison of models are present in most tools. Few frameworks accept custom evaluation criteria. Despite the fact that  many  frameworks include analysis processes that depend on custom properties, only 2 tools include an annotator GUI. Only 5 tools offer an interface for visualizing and filtering the Ground Truth (GT) data and  the model predictions (e.g., for filtering predictions by a specific  error type).
A final remark on the reporting capabilities: less than 30\% of the surveyed tools offer a way to build a comprehensive report with all the relevant metrics. Even less tools provide a summary organized  by class or by custom property.

\begin{table*}[h!]
\centering
\Large
\caption{Performance break down,  input distribution, custom properties and metrics, model comparison,  data editing and visualization functions.}
\label{tab:comp_feat}
\resizebox{\linewidth}{!}{
\begin{tabular}[b]{ l   K{0.95cm}  K{0.95cm}  K{0.95cm}  K{0.95cm}  K{0.95cm}  K{0.95cm}  K{0.95cm}  K{0.95cm}  K{0.95cm}  K{0.95cm}  K{0.95cm}  K{0.95cm}  K{0.95cm}  K{0.95cm}  K{0.95cm}  K{0.95cm}  K{0.95cm}  K{0.95cm}  K{0.95cm}  K{0.95cm}  K{0.95cm}  K{0.95cm}  K{0.95cm}  K{0.95cm}  K{0.95cm}  K{0.95cm}  K{0.95cm}  K{0.95cm}  K{0.95cm}  K{0.95cm} }
\hline
\textbf{Feature} & \textbf{\cite{dollar2009pedestrian}} & \textbf{\cite{hoiem2012diagnosing}} & \textbf{\cite{russakovsky2013detecting}} & \textbf{\cite{mscoco}} & \textbf{\cite{hariharan2014simultaneous}} & \textbf{\cite{zhu2015diagnosing}} & \textbf{\cite{amershi2015modeltracker}} & \textbf{\cite{redondo2016pose}} & \textbf{\cite{krause2016interacting}} & \textbf{\cite{zhang2016far}} & \textbf{\cite{ruggero2017benchmarking}} & \textbf{\cite{krause2017workflow}} & \textbf{\cite{ren2016squares}} & \textbf{\cite{sigurdsson2017actions}} & \textbf{\cite{alwassel2018diagnosing}} & \textbf{\cite{nekrasov2018diagnostics}} & \textbf{\cite{zhang2018manifold}} & \textbf{\cite{wexler2019if}} & \textbf{\cite{bolya2020tide}} & \textbf{\cite{torres2020odin}} & \textbf{\cite{padilla2020survey}} & \textbf{\cite{yoon2020tensorflow}} & \textbf{\cite{gleicher2020boxer}} & \textbf{\cite{demidovskij2021openvino}} & \textbf{\cite{padilla2021comparative}} & \textbf{\cite{Fan_2021_WACV}} & \textbf{\cite{chen2021diagnosing}} & \textbf{\cite{krater2021aideveloper}} & 
\textbf{\cite{nourani2022detoxer}}\\ \hline

\multicolumn{29}{c}{\textbf{Performance break down and input distribution  functions}} \\ \hline\hline

Overall Analysis & \textbf{\checkmark} & \textbf{\checkmark} & & \textbf{\checkmark} & \textbf{\checkmark} & \textbf{\checkmark} & \textbf{\checkmark} & \textbf{\checkmark} & \textbf{\checkmark} & \textbf{\checkmark} & \textbf{\checkmark} & \textbf{\checkmark} & & & \textbf{\checkmark} & \textbf{\checkmark} & & \textbf{\checkmark} & \textbf{\checkmark} & \textbf{\checkmark} & \textbf{\checkmark} & \textbf{\checkmark} & \textbf{\checkmark} & \textbf{\checkmark} & \textbf{\checkmark} & & \textbf{\checkmark} & & \textbf{\checkmark}\\ \hline
Per class Analysis & & \textbf{\checkmark} & & \textbf{\checkmark} & \textbf{\checkmark} & \textbf{\checkmark} & & \textbf{\checkmark} & & & & & \textbf{\checkmark} & \textbf{\checkmark} & & \textbf{\checkmark} & & \textbf{\checkmark} & & \textbf{\checkmark} & \textbf{\checkmark} & \textbf{\checkmark} & \textbf{\checkmark} & \textbf{\checkmark} & \textbf{\checkmark} & & & & \textbf{\checkmark}\\ \hline
Per property Analysis & \textbf{\checkmark} & \textbf{\checkmark} & \textbf{\checkmark} & \textbf{\checkmark} & & \textbf{\checkmark} & & \textbf{\checkmark} & & \textbf{\checkmark} & \textbf{\checkmark} & \textbf{\checkmark} & & \textbf{\checkmark} & \textbf{\checkmark} & \textbf{\checkmark} & & \textbf{\checkmark} & \textbf{\checkmark} & \textbf{\checkmark} & & \textbf{\checkmark} & \textbf{\checkmark} & & & \textbf{\checkmark} & \textbf{\checkmark} & &\\ \hline
Overall report & & & & \textbf{\checkmark} & & \textbf{\checkmark} & & & & & & & & & & & & & \textbf{\checkmark} & \textbf{\checkmark} & & \textbf{\checkmark} & \textbf{\checkmark} & & & & & \textbf{\checkmark} &\\ \hline
Per class report & & & & & & \textbf{\checkmark} & & & & & & & & & & & & & & \textbf{\checkmark} & & \textbf{\checkmark} & & & & & & &\\ \hline
Per property report & & & & \textbf{\checkmark} & & & & & & & & & & & & & & & & \textbf{\checkmark} & & & & & & & & &\\ \hline
Property distribution & & & & & & & & & & & & & & & & \textbf{\checkmark} & & \textbf{\checkmark} & & \textbf{\checkmark} & & & \textbf{\checkmark} & & \textbf{\checkmark} & \textbf{\checkmark} & & &\\ \hline
Class distribution & & & & & & & & & & & & & & & & & & \textbf{\checkmark} & & \textbf{\checkmark} & &  & \textbf{\checkmark} & & \textbf{\checkmark} & & &  \textbf{\checkmark} &\\ \hline

\multicolumn{29}{c}{\textbf{Custom properties and metrics}} \\ \hline\hline

Builtin input properties & \textbf{\checkmark} & \textbf{\checkmark} & \textbf{\checkmark} & & & \textbf{\checkmark} & & \textbf{\checkmark} & \textbf{\checkmark} & \textbf{\checkmark} & \textbf{\checkmark} & \textbf{\checkmark} & & \textbf{\checkmark} & \textbf{\checkmark} & & & \textbf{\checkmark} & \textbf{\checkmark} & \textbf{\checkmark} & & & \textbf{\checkmark} & & & \textbf{\checkmark} & \textbf{\checkmark} &\\ \hline
User-defined properties & & & & & & & & & \textbf{\checkmark} & & & \textbf{\checkmark} & & & \textbf{\checkmark} & & \textbf{\checkmark} & \textbf{\checkmark} & \textbf{\checkmark} & \textbf{\checkmark} & & & \textbf{\checkmark} & & & & & &\\ \hline
User-defined metrics & & & & & & & & & & & & & & & & & & & & \textbf{\checkmark} & & & & & & & & & \\ \hline

\multicolumn{29}{c}{\textbf{Editing, visualization and comparison functions}} \\ \hline\hline

Model comparison & \textbf{\checkmark} & & \textbf{\checkmark} & & & & & \textbf{\checkmark} & \textbf{\checkmark} & \textbf{\checkmark} & & & & \textbf{\checkmark} & & \textbf{\checkmark} & \textbf{\checkmark} & \textbf{\checkmark} & \textbf{\checkmark} & \textbf{\checkmark} & & \textbf{\checkmark} & \textbf{\checkmark} & & & \textbf{\checkmark} & & &\\ \hline

Annotator & & & & & & & & & & & & & & & & & & \textbf{\checkmark} & & \textbf{\checkmark} & & & & & & & & & \\ \hline
Visualizer & & & & & & & & & & & & & & & & & & & & \textbf{\checkmark} & \textbf{\checkmark} & \textbf{\checkmark} & & & & & & \textbf{\checkmark} & \textbf{\checkmark}\\ \hline
\end{tabular}
}
\end{table*}

\subsection{Beyond Computer Vision\label{sec:beyond}}

The focus of this survey is on the tools that assist  the CV tasks. The overview has been extended also to the frameworks that support the classification of other types of data, because they share most  metrics  and types of analysis with the surveyed image classification tools. However, the black-box diagnosis approach is relevant also in other scenarios in which DNNs are applied to such tasks  as time series analysis (TS),  natural language processing (NLP), and recommender systems (RS). In this section we  provide some essential references to the research and survey works that address the black-box diagnosis for tasks other than the CV ones.

A notable example is the application of DNNs to temporal data series  for such applications as anomaly detection  \citep{Pang2021anomaly,surveyanomaly} and predictive maintenance  \citep{surveymaintenance}. The time series data sets for such applications are characterized by many  properties (e.g., the sampling frequency, the stationarity and  periodicity of the series, the type and physical characteristics of the  signal and of the corresponding acquisition sensor). Such a richness of significant input properties could be exploited to enable the  break down of performance indicators and the attribution of errors to specific features of the input.   Recent works have started to implement off-the-shelf black-box error diagnosis and performance break down functionalities  \citep{Vollert2021predmaint}.   

Tools such as \citep{zoppi2019evaluation, herzen2021darts, carrasco2021anomaly} extend the support beyond the use of basic metrics in the evaluation phase and cover also the training and refinement of the model. The work  \citep{krokotsch2020novel} proposes novel temporal evaluation metrics and provides their implementation as command line scripts.  Several research and commercial frameworks assist the workflow of anomaly detection and predictive maintenance applications but do not  support  error attribution and metrics break down. An example is   RELOAD   \citep{zoppi2019evaluation},  which aids the ingestion of data,  the selection of the most informative
features,  the execution of multiple anomaly detection algorithms, the evaluation of alternative anomaly identification strategies, the evaluation of multiple metrics and the visualization of results in a GUI. RELOAD implements multiple metrics and algorithms off-the-shelf and has an extensible architecture. However it does not support yet the break down of performance metrics and the attribution of errors based on the features of the input. A black-box error diagnosis tool for time series is ODIN TS \citep{odints}, an extension of the ODIN  tool   \citep{torres2021odin} for time series analysis. ODIN TS  implements the  basic time series metrics and diagrams (accuracy, precision, recall, F1 score, miss alarm rate, false alarm rate, NAB score, MAE, MSE, Root Mean Squared Error, Mean Absolute Percentage Error, Precision-Recall and ROC curves), introduces new types of analysis for anomaly detection, such as FP error categorization,  enables the annotation of the time series and  the visualization of the data set and of the predictions.

In the NLP field, the BlackBox 
workshops series\footnote{\url{https://blackboxnlp.github.io/}} is dedicated to interpretability and diagnosis issues in the application of DNNs to  NLP problems, with a multidisciplinary approach spanning not only machine learning, but also psychology, linguistics, and neuroscience. As an example of this line of research, the  work \citep{gralinski2019geval} describes the GEval tool, a framework  for detecting anomalies in  NLP  test sets, supporting data preparation, detecting problems in the  model and comparing the output of alternative models.   LIT (Language Interpretability Tool) \citep{tenney2020language}, by Google Research,  is a framework focused on the interpretability of NLP models, which also supports  multiple evaluation metrics, counterfactual analysis and the visualization of the data set and of the model outputs. EXPATS (Explainable Automated Text Scoring) \citep{manabe2021expats} builds upon LIT and offers life cycle support for the specific NLP task of text scoring. It implements multiple feature extractors and metrics and can work with predefined or user-supplied models. The tool is open source and designed with an open architecture, to enable the plugin of custom analysis components.  

The proliferation of algorithms in the recommender system sector has spawned the interest for tools supporting the life cycle of system development, with the aim of standardizing the evaluation and  easing the reproduction of results.  
The recent contributions include  comprehensive libraries implementing the most popular RS algorithms, such as the RecBole library       
\citep{zhao2021recbole}, and tools supporting the  model development and assessment. A recent example of the available toolkits is ELLIOT \citep{anelli2021elliot} an open-source recommendation system development framework assisting pre-processing operations, enabling alternative hyperparameters optimization strategies and implementing multiple evaluation metrics. ELLIOT also covers bias and fairness analysis, supported by statistical significance
tests, a functionality particularly relevant in  recommendation scenarios.

 \section{Issues and research directions}\label{sec:issues}
    
Black-box error diagnosis is a viable complement to interpretability techniques for achieving a deeper understanding of the performances of DNNs. However, the panorama of the tools and frameworks that support error diagnosis shows that there are still margins for improvement and research before full maturity is attained.

\subsection{Open issues}

The analysis of the surveyed tools reveals several open issues.
\begin{itemize}
\item \textbf{Consensus}:  Average Precision and Accuracy are implemented off-the-shelf by most CV and classification tools.  But besides such basic metrics, there is little agreement among the tools addressing the same task about the core set of metrics and analyses that are most beneficial to performance and error diagnosis. Table \ref{tab:metricsxtask} shows task by task the percentages of tools that implement each metrics. 
\begin{table*}
\centering
\caption{The  tools that implement each metrics in percentage. Each column represents a task or a family of related tasks and shows the number of tools that support it.}
\resizebox{0.8\textwidth}{!}{
\begin{tabular}[c]{ ll c c c c c c }

\hline
\multicolumn{2}{ c }{Metrics-Analysis/Task} & CL (11) & OD/IS/SS (14) & PE (3) & AD (2) & OT (2) & VRD (1) \\ \hline
\multicolumn{1}{ l }{\multirow{26}{*}{Metrics}} & \multicolumn{1}{ p{5cm} }{Accuracy} & \textbf{73\% (8)} & 14\% (2) & 0\% (0) & - & - & - \\ \cline{2-8} 
\multicolumn{1}{ l }{} & \multicolumn{1}{ p{5cm} }{Error Rate} & 9\% (1) & - & - & - & - & - \\ \cline{2-8} 
\multicolumn{1}{ l }{} & \multicolumn{1}{ p{5cm} }{Precision} & 36\% (4) & 14\% (2) & 0\% (0) & 0\% (0) & 0\% (0) & 0\% (0) \\ \cline{2-8} 
\multicolumn{1}{ l }{} & \multicolumn{1}{ p{5cm} }{Recall} & 36\% (4) & 21\% (3) & 0\% (0) & 0\% (0) & 0\% (0) & 0\% (0) \\ \cline{2-8} 
\multicolumn{1}{ l }{} & \multicolumn{1}{ p{5cm} }{F1 score} & 36\% (4) & 14\% (2) & 0\% (0) & 0\% (0) & 0\% (0) & 0\% (0) \\ \cline{2-8} 
\multicolumn{1}{ l }{} & \multicolumn{1}{ p{5cm} }{Average Precision} & 18\% (2) & \textbf{64\% (9)} & \textbf{100\% (3)} & \textbf{100\% (2)} & \textbf{50\% (1)} & \textbf{100\% (1)} \\ \cline{2-8} 
\multicolumn{1}{ l }{} & \multicolumn{1}{ p{5cm} }{Average Recall} & 9\% (1) & 29\% (4) & 33\% (1) & 0\% (0) & \textbf{50\% (1)} & 0\% (0) \\ \cline{2-8} 
\multicolumn{1}{ l }{} & \multicolumn{1}{ p{5cm} }{ROC AUC} & 45\% (5) & - & - & - & - & - \\ \cline{2-8} 
\multicolumn{1}{ l }{} & \multicolumn{1}{ p{5cm} }{Precision-Recall AUC} & 36\% (4) & 7\% (1) & 0\% (0) & 0\% (0) & 0\% (0) & 0\% (0) \\ \cline{2-8} 
\multicolumn{1}{ l }{} & \multicolumn{1}{ p{5cm} }{F1 AUC} & 9\% (1) & 7\% (1) & 0\% (0) & 0\% (0) & 0\% (0) & 0\% (0) \\ \cline{2-8} 
\multicolumn{1}{ l }{} & \multicolumn{1}{ p{5cm} }{Mean Error} & 9\% (1) & - & - & - & - & - \\ \cline{2-8} 
\multicolumn{1}{ l }{} & \multicolumn{1}{ p{5cm} }{Mean Absolute Error} & 9\% (1) & - & - & - & - & - \\ \cline{2-8} 
\multicolumn{1}{ l }{} & \multicolumn{1}{ p{5cm} }{Mean Squared Error} & 9\% (1) & - & - & - & - & - \\ \cline{2-8} 
\multicolumn{1}{ l }{} & \multicolumn{1}{ p{5cm} }{Odds Ratio} & 9\% (1) & - & - & - & - & - \\ \cline{2-8} 
\multicolumn{1}{ l }{} & \multicolumn{1}{ p{5cm} }{Matthews Correlation Coefficient} & 18\% (2) & - & - & - & - & - \\ \cline{2-8} 
\multicolumn{1}{ l }{} & \multicolumn{1}{ p{5cm} }{Mean Intersection Over Union (Mean IoU)} & - & 21\% (3) & 0\% (0) & 0\% (0) & \textbf{100\% (2)} & 0\% (0) \\ \cline{2-8} 
\multicolumn{1}{ l }{} & \multicolumn{1}{ p{5cm} }{Miss Rate} & - & 21\% (3) & - & - & 0\% (0) & 0\% (0) \\ \cline{2-8} 
\multicolumn{1}{ l }{} & \multicolumn{1}{ p{5cm} }{Localization Latency} & - & 7\% (1) & - & 0\% (0) & 0\% (0) & 0\% (0) \\ \cline{2-8} 
\multicolumn{1}{ l }{} & \multicolumn{1}{ p{5cm} }{Success Score} & - & - & - & - & \textbf{50\% (1)} & - \\ \cline{2-8} 
\multicolumn{1}{ l }{} & \multicolumn{1}{ p{5cm} }{Failure Rate} & - & - & - & - & \textbf{50\% (1)} & - \\ \cline{2-8} 
\multicolumn{1}{ l }{} & \multicolumn{1}{ p{5cm} }{Spatio-Temporal Tube Average Precision (STT-AP)} & - & - & - & - & \textbf{50\% (1)} & - \\ \cline{2-8} 
\multicolumn{1}{ l }{} & \multicolumn{1}{ p{5cm} }{Average Viewpoint Precision} & - & - & 33\% (1) & - & - & - \\ \cline{2-8} 
\multicolumn{1}{ l }{} & \multicolumn{1}{ p{5cm} }{Average Orientation Similarity} & - & - & 33\% (1) & - & - & - \\ \cline{2-8} 
\multicolumn{1}{ l }{} & \multicolumn{1}{ p{5cm} }{Mean Angle Error} & - & - & 33\% (1) & - & - & - \\ \cline{2-8} 
\multicolumn{1}{ l }{} & \multicolumn{1}{ p{5cm} }{Median Angle Error} & - & - & 33\% (1) & - & - & - \\ \cline{2-8} 
\multicolumn{1}{ l }{} & \multicolumn{1}{ p{5cm} }{Object Keypoint Similarity} & - & - & 33\% (1) & - & - & - \\ \hline
\multicolumn{1}{ c }{\multirow{3}{*}{Curves}} & \multicolumn{1}{ p{5cm} }{ROC curve} & 36\% (4) & - & - & - & - & - \\ \cline{2-8} 
\multicolumn{1}{ c }{} & \multicolumn{1}{ p{5cm} }{Precision-Recall curve} & 36\% (4) & 14\% (2) & 33\% (1) & 0\% (0) & 0\% (0) & 0\% (0) \\ \cline{2-8} 
\multicolumn{1}{ c }{} & \multicolumn{1}{ p{5cm} }{F1 curve} & 10\% (1) & 7\% (1) & 0\% (0) & 0\% (0) & 0\% (0) & 0\% (0) \\ \hline
\multicolumn{1}{ c }{\multirow{16}{*}{Analysis}} & \multicolumn{1}{ p{5cm} }{Confusion Matrix} & \textbf{73\% (8)} & - & - & - & - & - \\ \cline{2-8} 
\multicolumn{1}{ c }{} & \multicolumn{1}{ p{5cm} }{\# False Positives (FP)} & 27\% (3) & 14\% (2) & 0\% (0) & 0\% (0) & 0\% (0) & 0\% (0) \\ \cline{2-8} 
\multicolumn{1}{ c }{} & \multicolumn{1}{ p{5cm} }{\# False Negatives (FN)} & 27\% (3) & 14\% (2) & 0\% (0) & 0\% (0) & 0\% (0) & 0\% (0) \\ \cline{2-8} 
\multicolumn{1}{ c }{} & \multicolumn{1}{ p{5cm} }{True Positive Analysis} & 9\% (1) & 7\% (1) & 0\% (0) & 0\% (0) & 0\% (0) & 0\% (0) \\ \cline{2-8} 
\multicolumn{1}{ c }{} & \multicolumn{1}{ p{5cm} }{False Positive Analysis} & 9\% (1) & 50\% (7) & \textbf{100\% (3)} & \textbf{100\% (2)} & 50\% (1) & \textbf{100\% (1)} \\ \cline{2-8} 
\multicolumn{1}{ c }{} & \multicolumn{1}{ p{5cm} }{False Negative Analysis} & 9\% (1) & 29\% (4) & 66\% (2) & 50\% (1) & 0\% (0) & \textbf{100\% (1)} \\ \cline{2-8} 
\multicolumn{1}{ c }{} & \multicolumn{1}{ p{5cm} }{True Negative Analysis} & 9\% (1) & - & - & - & - & - \\ \cline{2-8} 
\multicolumn{1}{ c }{} & \multicolumn{1}{ p{5cm} }{Error Categorization} & 9\% (1) & 43\% (6) & \textbf{100\% (3)} & \textbf{100\% (2)} & 50\% (1) & \textbf{100\% (1)} \\ \cline{2-8} 
\multicolumn{1}{ c }{} & \multicolumn{1}{ p{5cm} }{False Positive Error Categorization} & 9\% (1) & 43\% (6) & 100\% (3) & 100\% (2) & 50\% (1) & \textbf{100\% (1)} \\ \cline{2-8} 
\multicolumn{1}{ c }{} & \multicolumn{1}{ p{5cm} }{False Negative Error Categorization} & 9\% (1) & 14\% (2) & 0\% () & 0\% () & 0\% () & \textbf{100\% (1)} \\ \cline{2-8} 
\multicolumn{1}{ c }{} & \multicolumn{1}{ p{5cm} }{Error Contribution Analysis} & 9\% (1) & 50\% (7) & 66\% (2) & 50\% (1) & 0\% (0) & \textbf{100\% (1)} \\ \cline{2-8} 
\multicolumn{1}{ c }{} & \multicolumn{1}{ p{5cm} }{Intersection Over Union Analysis} & - & 36\% (5) & 33\% (1) & 0\% (0) & \textbf{50\% (1)} & 0\% (0) \\ \cline{2-8} 
\multicolumn{1}{ c }{} & \multicolumn{1}{ p{5cm} }{Reliability Analysis} & 9\% (1) & 7\% (1) & 0\% (0) & 0\% (0) & \textbf{50\% (1)} & 0\% (0) \\ \cline{2-8} 
\multicolumn{1}{ c }{} & \multicolumn{1}{ p{5cm} }{Temporal Reasoning} & - & - & - & \textbf{50\% (1)} & 0\% (0) & - \\ \cline{2-8} 
\multicolumn{1}{ c }{} & \multicolumn{1}{ p{5cm} }{Person-based Reasoning} & - & - & - & \textbf{50\% (1)} & 0\% (0) & - \\ \cline{2-8} 
\multicolumn{1}{ c }{} & \multicolumn{1}{ p{5cm} }{Qualitative Analysis (Visualizer)} & 27\% (3) & 21\% (3) & 0\% (0) & - & 0\% (0) & 0\% (0) \\ \hline
\end{tabular}
}
\label{tab:metricsxtask}
\end{table*} A common effort to define a core set of metrics per task must go beyond the few metrics popularized by the benchmarks focused on end-to-end evaluation. The definition of a consensus set of metrics and analyses  would promote the design of methodological guidelines for black-box error diagnosis based on  fundamental performance indicators and diagnosis reports available off-the-shelf.  It would also help model and framework developers save time in the implementation of diagnosis metrics and reports and concentrate on more advanced functionalities of their architectures and tools.  The starting point for the definition of a consensus set of indicators is the abundant literature on the performance metrics most suitable for each task  \citep{monteiro2006performance,hossin2015review,novakovic2017evaluation,padilla2020survey, padilla2021comparative}.

\item  \textbf{Workflow coverage}: all the analyzed tools focus on prediction performance evaluation with few considering also the other life cycle phases of Figure \ref{fig:MLdevproc.drawio}. Thus, a complete development workflow would require the use of multiple tools, each one with its own input/output formats, configuration, metrics, and visualizations/reports. Integrating in one solution the support to all the life cycle phases would accelerate the model development, evaluation and  refinement loop. This requirements is especially relevant to those tools that offer the break down of metrics based on custom properties, which should be added manually to the input data set, or extracted automatically from it, during the preparation phase. 
\item  \textbf{Visual analytics and support for qualitative analysis}: the surveyed tools include almost only quantitative metrics or analyses that are useful to measure model performance or diagnose errors. In particular, when dealing with visual data, these measures are not enough to fully understand the model behavior. Qualitative analysis is fundamental to visualize both errors and correct predictions and, associated with interpretability techniques, may help discover unwanted correlations. The interfaces of error diagnosis tools could be extended to support the automatic visualization of data samples relevant for qualitative analysis, e.g., those falling into given ranges of one or more output metrics or those displaying a certain type of error. 

\item \textbf{Automatic extraction of properties}: custom properties associated to the input data have been shown to be beneficial in diagnosing and categorizing errors. Such properties in part can be automatically derived from the data (e.g., image color space, bounding box size and aspect ratio, difficulty and quality level, number of objects, etc.).  Some of the analyzed tools compute only elementary properties (e.g., bounding box size or image color scheme) and none integrate an approach to extract non trivial diagnosis-oriented properties automatically.

\item \textbf{Data quality assessment}:  errors that are attributed to the model may be due to the presence of noise in the ground truth annotations. Only few tools integrate data quality and  error diagnosis and provide head to head comparison of the same model on different data sets. A particular case of data quality analysis is bias detection in the input data set, which is the subject of a completely distinct family of tools  \citep{mehrabi2021survey} but is supported only by one black-box error diagnosis framework   \citep{wexler2019if}.  

\item \textbf{Scalability}:  the authors of several tools openly admit that their framework cannot be applied to very large data sets, featuring many samples, many classes, or many custom properties. The difficulty stems from both the computational effort required (e.g., metrics may be computationally heavy) and from the visualization of results (traditional plots,  interactive views, and summaries get cluttered and difficult to read).

\item \textbf{Architecture and API standardization}: Albeit most tools publish their implementation code in open source repositories, the lack of a common  architecture and of standard APIs makes the implementation of metrics, property extractors and visualizations/reports non portable and requires the re-implementation or the wrapping of even the most basic performance indicators. The definition of a plug-in architecture and of standard module interfaces is typical in more  mature fields such as software development, as witnessed e.g., by the popular Eclipse\footnote{ \url{https://www.eclipse.org}} and JetBrains\footnote{ \url{https://www.jetbrains.com/}} frameworks.
A similar approach applied to black-box diagnosis tools and more generally to DNN workflow management tools would promote the development of a community-managed library of reusable metrics,  property extractors and visualizations/reports which could be installed rather than re-implemented in any given framework.
\end{itemize}

\subsection{Research directions}
The review of DNNs black-box diagnosis frameworks uncovers many research challenges that are still to be pursued.
\begin{itemize}
\item  \textbf{Integration of black-box diagnosis and interpretability techniques}: interpretability in machine learning is defined as \quoted{the ability to explain or to present in understandable terms to a human}  \citep{doshi2017towards,GuidottiMRTGP19}  how a model makes a prediction. The interpretability of DNNs  has attracted increasing attention by researchers  \citep{MONTAVON20181} due to the impact that this class of algorithms has in critical domains such as medicine, economy, safety and social sciences. Interpretability techniques offer a view of the system behavior alternative but closely related to that afforded by black-box error diagnosis. They seek to unveil \textit{interpretability factors} i.e., human-understandable concepts and processes that are at the base of the model prediction. Black-box error diagnosis and performance break down can aid the discovery of interpretability factors: if a model consistently fails or succeeds when the input exhibits certain human-defined properties, this is a hint that such properties play a role in the interpretability. For example, if a system for iconography classification in art images  \citep{milani2021dataset}  consistently classifies representations of a given subject (e.g., images of Saint Jerome in Christian art paintings) with more accuracy and confidence when the input contains distinctive symbols (e.g., a lion couched at the saint's feet or the cardinal's galero) this finding could attest the interpretative value of such symbols in the classification of that character. The relationship between interpretability and black-box analysis is exploited in methods, such as  \citep{petsiuk2020black}, that aim at discovering interpretability factors  by modifying the input data and then measuring the changes in the output (e.g., by suppressing some features or masking part of an image), and \cite{theissler2022confusionvis,theissler2020ml}, which allow the user to understand and interpret the models outputs based on the confusion matrices.

\item \textbf{Integration of performance break down and impact analysis with saliency/attention maps:} a specific case of integration of interpretability techniques and black-box analysis occurs in CV:  DNNs interpretability approaches compute saliency or attention maps which highlight  the pixels that are more important for the classification of the image. The saliency/attention maps could help the designer discover some properties of the input that affect  prediction performance. Such properties could then be added as annotations to the input data set and used by the performance break down functionality to quantify their impact on the prediction performances.  A step in  this direction is the SECA system  \citep{BalaynSL0B21}, which uses crowdsourced conceptual labels associated to saliency maps to enable explanation queries about the inference made by DNNs for image classification. The interpretability factors derived in this way could be exploited for error diagnosis and performance break down bridging the gap between performance-oriented and interpretation-oriented model evaluation.

\item  \textbf{Integration of error diagnosis and runtime performance analysis}: at the beginning of their diffusion DL models were designed to obtain the best possible performance on GPU-accelerated systems. With the emergence of mobile and edge AI applications the portability of DL models to constrained hardware (e.g., mobile phones and embedded devices) has become a major research topic  \citep{ChenZZ0S020}. The profiling of models (e.g., with metrics based on throughput, occupied memory, latency, or operation level) helps optimize architectures and support math-limited or memory-limited devices. Some techniques have been proposed, such as model pruning and quantization, but their automatic application and the parameters optimization search are still ongoing research. Only one tool  \citep{demidovskij2021openvino} among the surveyed ones integrates runtime performance analysis and some elementary diagnosis functions such as accuracy checking and model comparison. 
Given that in hardware-constrained scenarios the trade off between accuracy and runtime performances is a prominent concern, integrating the two perspectives would produce a constraint-aware tool extending error diagnosis and performance break down to scenarios with hardware limitations.  
 
\item  \textbf{Model design guidance}: deep neural models are not mere data driven tools. They embody a lot of prior knowledge expressed implicitly in the structure of the architecture, in the selection of the operators and in the definition of the training strategy and of the data set. 
A recent research field, the so called automated deep learning (AutoDL), addresses the \textit{combined selection and hyperparameter optimization of classification algorithms}   \citep{Talbi2021autodl,thornton2013auto}. AutoDL research investigates the design patterns that can be applied to support or even minimize the human effort in the definition of optimal DL models for a variety of tasks  \citep{liu2020towards,dong2021automated}.  Current methods mostly rely on neural architecture search, which applies a rather brute force space exploration approach to the distillation of the best model for a given task.
An extremely interesting evolution of the future error diagnosis frameworks would be to turn them into tools capable not only of diagnosing problems, but also to recommend model improvements. This would require a mapping between the model weaknesses identified by error diagnosis and a portfolio of model refactoring and improvement operators distilling the current wisdom of manual and automated DL design. 
Such a progress would somehow reconcile the  practices of software development and data driven model design, which are now regarded as completely secluded and move the AutoDL field beyond the mere architecture search.     

\item \textbf{New task and data types}: As shown in Table \ref{tab:tools} and in the brief overview of Section \ref{sec:beyond}, the surveyed frameworks focus mostly on classification and CV localization tasks. The black-box  error diagnosis approach and tools could benefit also other domains especially those featuring complex data  and non trivial performance indicators. 
\end{itemize} \section{Conclusions}\label{sec:concl}
This survey presented the existing tools for the black-box diagnosis of errors in DL models for CV tasks. Major properties that would guide the user choice have been discussed and analyzed: supported tasks and media types, implemented metrics and analyses, performance break down functionalities, customization capabilities and openness. Novelties, advantages and disadvantages of the surveyed works have been described to provide the reader with a clear and up to date view of the field. Metrics, analyses and functionalities have been collected from each work and grouped  by task to ease the comparison between the tools with a common focus. Several issues emerged  from the survey have been discussed and the most promising research directions have been highlighted to help improve the present state of the art.

\backmatter
\bmhead{Acknowledgments}

This work is partially supported by the project \quoted{PRECEPT - A novel decentralized edge-enabled PREsCriptivE and ProacTive framework for increased energy efficiency and well-being in residential buildings} funded by the EU H2020 Programme, grant agreement no. 958284.

\section*{Declarations}

\textbf{Conflict of interest.} The authors declare that the research was conducted
in the absence of any commercial or financial relationships that could
be construed as a potential conflict of interest.\\
\textbf{Data availability.} Data sharing not applicable to this article as no datasets were generated or analysed during the current study

\section*{List of Abbreviations}
\textbf{AD}\tab Action Detection\\ 
\textbf{AI}\tab Artificial Intelligence\\ 
\textbf{AP}\tab  Average Precision\\ 
\textbf{AUC}\tab  Area Under the Curve\\ 
\textbf{CAM}\tab  Class Activation Map\\ 
\textbf{CL}\tab  Classification\\ 
\textbf{CV} \tab Computer Vision\\ 
\textbf{DNN}\tab  Deep Neural Network\\ 
\textbf{ET}\tab  Error Type\\ 
\textbf{FN} \tab False Negative\\ 
\textbf{FP}\tab  False Positive\\ 
\textbf{GT}\tab  Ground Truth\\ 
\textbf{IoU}\tab  Intersection over Union\\ 
\textbf{IS}\tab  Instance Segmenttion\\ 
\textbf{MAE} \tab Mean Absolute Error\\ 
\textbf{mAP}\tab  mean Average Precision\\ 
\textbf{ME} \tab Mean Error\\ 
\textbf{ML}\tab  Machine Learning\\ 
\textbf{MSE}\tab  Mean Squared Error\\ 
\textbf{NAB}\tab  Numenta Anomaly Benchmark\\ 
\textbf{NLP}\tab  Natural Language Processing\\ 
\textbf{OD} \tab Object Detection\\ 
\textbf{OT}\tab  Object Tracking\\ 
\textbf{PE}\tab  Pose Estimation\\ 
\textbf{PR} \tab Precision-Recall\\ 
\textbf{RMSE}\tab Root Mean Squared Error\\ 
\textbf{ROC}\tab  Receiver Operating Characteristic\\ 
\textbf{RS} \tab  Recommender Systems\\ 
\textbf{SS} \tab Semantic Segmentation\\ 
\textbf{TN} \tab True Negative\\ 
\textbf{TP} \tab True Positive\\ 
\textbf{TS} \tab Time Series\\ 
\textbf{VRD} \tab Video Relation Detection

\bibliographystyle{sn-basic}

\bibliography{biblio}

\end{document}